\documentclass{article}

\usepackage{microtype}
\usepackage{graphicx}
\usepackage{subfigure}
\usepackage{xfp}
\usepackage{booktabs} %
\usepackage{placeins}
\usepackage{enumitem}

\usepackage{hyperref}

\usepackage[accepted]{icml2026}

\usepackage{amsmath}
\usepackage{amssymb}
\usepackage{mathtools}
\usepackage{amsthm}

\usepackage{comment}
\usepackage{color}
\usepackage{xcolor}
\usepackage{tikz}
\usepackage{hyperref}
\usepackage{graphicx}
\usepackage{siunitx}
\usepackage{threeparttable}
\usepackage{multicol}
\usepackage{multirow}
\usepackage{wrapfig}

\usepackage{xspace}
\makeatletter
\DeclareRobustCommand\onedot{\futurelet\@let@token\@onedot}
\def\@onedot{\ifx\@let@token.\else.\null\fi\xspace}

\def\eg{{e.g}\onedot} 
\def\ie{{i.e}\onedot}

\makeatother
\usepackage[capitalize,noabbrev]{cleveref}

\theoremstyle{plain}

\theoremstyle{definition}

\theoremstyle{remark}

\usepackage[textsize=tiny]{todonotes}

\icmltitlerunning{Differentiable Weightless Controllers: Learning Logic Circuits for Continuous Control}

\newcommand{\acronym}{DWC\xspace}
\newcommand{\acronyms}{DWCs\xspace}

\newcommand{\methods}{Differentiable Weightless Controllers\xspace}

\definecolor{istaDark}{HTML}{005844}

\definecolor{istaDark}{HTML}{005844}
\definecolor{istaMid}{HTML}{2B8A70} 

\usepackage{pgfplots}
\usepgfplotslibrary{external}
\usepgfplotslibrary{statistics,colorbrewer}
\pgfplotsset{compat=newest, cycle list/Set1-7, }
\usepgfplotslibrary{groupplots,fillbetween}
\usepackage{pgfplotstable}
 \usepgfplotslibrary{groupplots}
\pgfplotsset{
  myaxis/.style={
    width=5.cm, height=4.0cm,
    grid=both,
    xlabel={},
    title style={font=\small},
    label style={font=\small},
    tick label style={font=\small},
    legend style={font=\footnotesize},
    line width=0.9pt,
    tick style={line width=0.6pt},
    scaled ticks=false,
    xmin=0, xmax=1000000,
    xtick={0,500000,1000000},
    xticklabels={0,0.5M,1M},
  },
  meanline/.style={very thick, no markers},
  fpstyle/.style={densely dashed, very thick, no markers, draw=blue},
  fpband/.style={draw=blue, fill=blue, opacity=0.15},
  bandfill/.style={draw=none, opacity=0.15},
  base bar style/.style={
        ybar,
        ymin=0,
        ylabel style={align=center},
        legend style={at={(0.5,-0.2)}, anchor=north},
        error bars/y dir=both,
        error bars/y explicit,
        /pgf/bar width=1, %
    },
}
\pgfplotscreateplotcyclelist{myBoxplotColors}{
  {fill=blue!70!black},
  {fill=orange!85!black},
  {fill=red!75!black},
  {fill=violet!80!black},
  {fill=teal!70!black},
  {fill=magenta!70!black},
  {fill=gray!70!black},
  {fill=brown!70!black}
}
\pgfplotscreateplotcyclelist{myBoxplotColors6}{
  {fill=blue!70!black},
  {fill=orange!85!black},
  {fill=red!75!black},
  {fill=violet!80!black},
  {fill=teal!70!black},
  {fill=magenta!70!black},
  {fill=gray!70!black}
}
\pgfplotscreateplotcyclelist{myBoxplotColors5}{
  {fill=blue!70!black},
  {fill=orange!85!black},
  {fill=red!75!black},
  {fill=violet!80!black},
  {fill=teal!70!black}
}

\newcommand{\thermbarV}[4]{%
  \def\step{0.16} %
  \pgfmathsetmacro\half{0.5*(#4-1)*\step} %
  \coordinate (#2-base) at ($(#1.east)+(3mm,\half cm)$);
  \foreach \i [evaluate=\i as \dy using {\step*(\i-1)}] in {1,...,#4} {%
    \path (#2-base) ++(0,-\dy cm)
      node[bit,anchor=west] (t#2-\i) {};
    \ifnum\i>#3\relax\else
      \fill (t#2-\i.south west) rectangle (t#2-\i.north east);
    \fi
  }%
  \coordinate (#2-mid) at ($(t#2-1)!0.5!(t#2-#4)$);
}

\pgfplotsset{
  myaxis/.style={
    width=7.2cm, height=4.2cm,
    xmin=2, xmax=8,
    xtick={8,7,6,5,4,3,2},
    x dir=reverse, 
    scaled y ticks=false,
    xlabel={}, ylabel={Reward},
    ymajorgrids=true,
    legend cell align={left},
    error bars/y explicit,
    error bars/error bar style={line width=0.7pt},
    error bars/error mark options={line width=0.7pt, rotate=90, mark size=1.8pt},
  },
  box legend/.style={
            area legend, 
            draw=black, 
            fill opacity=0.6, 
            thick, 
            fill=#1,
  },
  sizeaxis/.style={
    width=3.4cm, height=3.4cm,
    x dir=reverse, 
    xlabel={}, ylabel={Reward},
    ymajorgrids=true,
    legend cell align={left},
    error bars/y explicit,
    error bars/error bar style={line width=0.7pt},
    error bars/error mark options={line width=0.7pt, rotate=90, mark size=1.8pt},
  },
  stdaxis/.style={
    width=3.4cm, height=3.4cm, 
    xlabel={}, 
    ylabel={Reward [$\times 1000$]},
    yticklabel={\pgfmathparse{\tick/1000}\pgfmathprintnumber{\pgfmathresult}}, ymajorgrids=true,
    error bars/y explicit,
    error bars/error bar style={line width=0.7pt},
    error bars/error mark options={line width=0.7pt, rotate=90, mark size=1.8pt},
  },
timeaxis/.style={
    width=3.5cm, height=3.5cm,
    grid=both,
    xlabel={},
    title style={font=\small},
    label style={font=\small},
    scaled y ticks=false,
    legend style={font=\footnotesize},
    line width=0.9pt,
    tick style={line width=0.6pt},
    scaled ticks=false,
    xmin=0, xmax=1000000,
    xtick={0,500000,1000000},
    xticklabels={0,0.5M,1M},
  },
  fpstyle/.style={line width=1.4pt, dashed, color=black!55},
  fpband/.style={draw=none, fill opacity=0.15},
     qA/.style={
      line width=1.2pt,
      color={rgb,1:red,0.1216;green,0.4667;blue,0.7059}, %
      mark=*,
      mark size=2.0pt,
      mark options={solid},     %
    },
    qB/.style={
      line width=1.2pt,
      color={rgb,1:red,1.0;green,0.498;blue,0.0549},      %
      mark=triangle*,
      mark size=2.2pt,
      mark options={solid},
    },
    qC/.style={
      line width=1.2pt,
      color={rgb,1:red,0.1725;green,0.1725;blue,0.6275},  %
      mark=square*,
      mark size=2.0pt,
      mark options={solid},
    },
    qD/.style={
      line width=1.2pt,
      color={rgb,1:red,0.6275;green,0.1;blue,0.1},  %
      mark=star,
      mark size=3.0pt,
      mark options={solid},
    },
    qAN/.style={
      line width=1.2pt,
      color={rgb,1:red,0.1216;green,0.4667;blue,0.7059}, %
      mark=None,
      mark size=2.0pt,
      mark options={solid},     %
    },
    qBN/.style={
      line width=1.2pt,
      color={rgb,1:red,1.0;green,0.498;blue,0.0549},      %
      mark=None,
      mark size=2.2pt,
      mark options={solid},
    },
    qCN/.style={
      line width=1.2pt,
      color={rgb,1:red,0.1725;green,0.6275;blue,0.1725},  %
      mark=None,
      mark size=2.0pt,
      mark options={solid},
    },
}
\tikzset{
  fpband/.style={fill opacity=0.15, draw=none, fill=black},
  qAband/.style={fill opacity=0.15, draw=none, fill=blue},
  qBband/.style={fill opacity=0.15, draw=none, fill=orange},
  qCband/.style={fill opacity=0.15, draw=none, fill=blue},
}

\definecolor{sz512}{RGB}{217,95,2}   %

\usetikzlibrary{arrows.meta,positioning}
\usetikzlibrary{positioning,arrows.meta,calc,fit,backgrounds,shapes.misc}

\title{\methods: \\Learning Logic Circuits for Continuous Control}
\makeatletter
\AtBeginDocument{%
  \@ifundefined{tikzifexternalizing}{}{%
    \tikzifexternalizing{\ClearShipoutPicture}{}
  }%
}
\makeatother

\newcommand{\myparagraph}[1]{\noindent\textbf{#1}\xspace} %

\begin{document}
  
\setlength{\dbltextfloatsep}{4pt plus 1.0pt minus 2.0pt}
\setlength{\dblfloatsep}{4pt plus 1.0pt minus 2.0pt}
\addtolength{\abovedisplayskip}{-.15\baselineskip}
\addtolength{\belowdisplayskip}{-.15\baselineskip}

\twocolumn[
\icmltitle{\methods: \\Learning Logic Circuits for Continuous Control}

\icmlsetsymbol{equal}{*}

\begin{icmlauthorlist}
\icmlauthor{Fabian Kresse}{ista}
\icmlauthor{Christoph H. Lampert}{ista}
\end{icmlauthorlist}

\icmlaffiliation{ista}{ISTA (Institute of Science and Technology Austria), 3400 Klosterneuburg, Austria}

\icmlcorrespondingauthor{Fabian Kresse}{fabian.kresse@ist.ac.at}

\icmlkeywords{Machine Learning, ICML}

\vskip 0.3in
]

\printAffiliationsAndNotice{}  %

\begin{abstract}
Controlling autonomous systems under real-world conditions 
often requires policies that can be evaluated with low 
latency and minimal energy consumption. 
Unfortunately, these conditions are at odds with the use 
of high-precision deep neural networks as controllers.  
In this work, we introduce \methods (\acronyms), a symbolic-differentiable 
architecture that learns flexible, non-linear, yet 
highly efficient control policies.
\acronyms can be trained end-to-end via gradient-based techniques, 
yet compile directly into FPGA-compatible circuits with 
few- or even single-clock-cycle latency and nanojoule-level 
energy cost per action.
Across five MuJoCo benchmarks, including high-dimensional Humanoid, 
\acronyms achieve returns competitive with standard deep %
policies (full-precision or quantized neural networks).
Furthermore, \acronyms exhibit structurally sparse and interpretable 
connectivity patterns, enabling direct inspection of which 
input values influence control decisions. 

\end{abstract}

\section{Introduction}

Deep Learning has transformed countless fields, from natural language processing \citep{vaswani2017attention,radford2019language} and computer vision \citep{krizhevsky2012imagenet,dosovitskiy2020image} to game playing \citep{mnih2013playing,silver2016mastering}. 
The paradigm has been successfully transferred to continuous control as \emph{deep reinforcement learning} (RL), where neural network {policies} serve as real-valued function approximators, trained to solve highly complex tasks,
such as quadrotor racing~\citep{kaufmann2023champion}, robot locomotion~\citep{miller2025high}, and even fusion-plasma control~\citep{degrave2022magnetic}.

\begin{figure}[t]
\hspace{-0.6cm}

\includegraphics{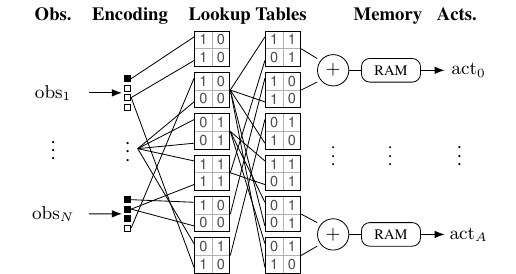}
\vspace{-0.3cm}

\caption{\methods (\acronyms): real-valued observations are thermometer-encoded into bitvectors, processed by two layers of multi-input boolean-output lookup tables (here drawn with 2 inputs), aggregated by group summation, and mapped via per-action memory lookups to final action values.}
\label{fig:bccs}
\end{figure}

However, standard neural networks rely on large numbers of compute-intense multiply and 
accumulate operations, which makes them difficult to run efficiently on resource-constrained 
platforms, such as UAVs and mobile robots.
For such cases, policies implemented as small and discrete-valued functions are 
preferable, which allow efficient implementation on embedded hardware,
such as FPGAs. 

Even on high-end GPUs, low-precision floating point or integer-only operations have 
efficiency advantages. Therefore, several prior works have studied either converting 
standard-trained deep networks into a \emph{quantized} representation (post-training 
compressions, PTQ), or learning deep networks directly in quantized form 
(quantization-aware training, QAT), see \eg, \citet{gholami2022survey} for a survey.
However, most such works target a static supervised or next-word-prediction setting, 
in which the input and output domains are discrete to start with, 
and the task involves predicting individual outputs for individual inputs.

The setup of \emph{continuous control}, \ie learning control policies for 
cyber-physical systems such as autonomous robots or wearable devices, is more 
challenging, because input \emph{states} and output \emph{actions} can be 
continuous-valued instead of categorical, and the controller is meant to be 
run repeatedly over long durations, such that even small errors in the 
control signal might accumulate over time.

Consequently, only few methods for quantized policy learning have been proposed 
so far. 
\citet{krishnan2019quarl} discuss PTQ from a perspective of reducing resources during training in RL and, in QAT ablations, find that low-bit quantization often does not harm returns. \citet{ivanov2025neural} investigate both QAT and pruning (\ie, reducing the number of weights) jointly, finding that the combination of high sparsity and 8-bit weight quantization does not reduce control performance.  
Very recently, \citet{kresse2025learningquantizedcontinuouscontrollers} 
showed that controllers can be trained using QAT with neurons that have 
weights and internal activation values of only 2 or 3 bits; we evaluate our approach against this baseline in the experiment section. 

In this work, we challenge the necessity of relying on \emph{weight-based} neural networks, \ie, those that rely on large matrix multiplications during inference, for learning continuous control policies. 
Instead, we introduce \emph{\methods (\acronyms)}, a symbolic-differentiable 
architecture for continuous control in which dense matrix multiplication 
is replaced with sparse boolean logic.
Figure~\ref{fig:bccs} illustrates the inference pipeline.

At their core lie \emph{differentiable weightless networks (DWNs)}~\citep{bacellar2024differentiable}, 
an FPGA-compatible and alternative variant of \emph{logic-gate networks (LGNs)}~\citep{petersen2022deep}
that were originally proposed for high-throughput, low-energy classification tasks. 
We extend the DWN architecture to the continuous RL domain, enabling synthesis 
of control policies that are digital circuits instead of real-valued functions. 
Specifically, we introduce an adaptive input encoding for converting real-valued 
input signals into binary vectors, and a trainable output decoding that converts 
binary output vectors into real-valued actions of suitable range and scale. 
The resulting \acronyms are compatible with gradient-based RL, and we demonstrate 
experimentally that the learned \acronyms match standard weight-based neural network 
baselines (full-precision or quantized) on MuJoCo benchmarks. %

At deployment time, \acronyms process continuous observations by quantizing 
them using {quantile binning} and thermometer encoding, resulting in 
a fixed-width bitvector representation for each observation dimension. 
The bitvectors are concatenated and processed by %
sparsely-connected layers of boolean-output lookup tables (LUTs)~\citep{bacellar2024differentiable}. 
The final layer produces a fixed number of bit outputs per action dimension.
These are summed using a \emph{popcount} operation and converted to a final 
action value using a single-cycle (SRAM) memory lookup. 

As a consequence, \acronyms are compatible with embedded hardware platforms, 
especially FPGAs, which explicitly support LUT operations. 
Here, \acronyms can run with few- or even single-cycle latency and minuscule 
(\eg Nanojoule-level) energy per operation, as we demonstrate for the case 
of an AMD Xilinx Artix-7.

Besides their efficiency, \acronyms also offer a potential gain over standard
networks in terms of their interpretability, because they consist of sparse, 
discrete, and symbolic elements instead of dense, continuous, matrix-multiplications 
in standard networks.
For instance, the sparse connectivity in the first layer allows for direct identification of the input dimensions and thresholds utilized by the controller.

\myparagraph{Contributions.}
To summarize, \textbf{our main contribution is the symbolic-differentiable \acronym architecture} 
that extends previous DWNs from classification to continuous control tasks.
We demonstrate that
\begin{itemize}[nosep] %
\smallskip\item \textbf{\acronyms can be trained successfully using standard RL algorithms}, reaching 
parity with floating-point policies (full-precision or quantized) in most of our experiments.

\smallskip\item \textbf{\acronyms are orders of magnitude more efficient than previous architectures}, achieving few- or even single-clock-cycle latency and nanojoule-level energy cost per action when 
compiled to FPGA hardware.

\smallskip\item \textbf{\acronyms allow for straightforward interpretation} of some aspects of the policies they implement, specifically which input dimensions matter for the decisions, and what the relevant 
thresholds are.
\end{itemize}

\section{Background}

\label{sec:background}

We briefly review deep reinforcement learning for 
continuous control, and provide background details on DWNs. %

\subsection{Reinforcement Learning (RL)} 
Reinforcement learning studies how an agent can learn, through trial and error, 
to maximize \emph{return} (cumulative reward) while interacting with an 
environment \cite{barto2021reinforcement}.  
There exist various reinforcement learning algorithms that are compatible with our 
setting of continuous control (continuous actions and observations). In the main 
body of this work we investigate the performance of \acronyms with the state-of-the-art 
\emph{Soft Actor-Critic (SAC)} method~\citep{haarnoja2018soft}. 
Results for \emph{Deep Deterministic Policy Gradient (DDPG)}~\citep{lillicrap2015continuous} 
and \emph{Proximal Policy Optimization (PPO)}~\citep{schulman2017proximal} can be found in Appendix \ref{app:ddpgppo}. 

SAC is an off-policy method that keeps a buffer of previous state transitions and taken actions. 
It updates the parameters of the policy network based on this buffer with soft-Q value estimations 
from two auxiliary networks (called \emph{critics}). 
Actions during training are sampled stochastically from a normal distribution, parametrized 
as $\mathcal{N}(\mu_\theta, \sigma_\theta)$, with $\theta$ being learned parameters. 
During deployment, the mean action is used deterministically.

\subsection{Differentiable Weightless Neural Networks}
Weightless networks~\cite{aleksander1984wisard,ludermir1994weightless} 
rely on table lookups instead of arithmetic operations (specifically, matrix multiplications): 
each neuron is a lookup table (LUT) with $k$ binary inputs and a single-bit output. 
Because each LUT input is connected to precisely one of the preceding layer outputs, 
this structure results in a sparsely connected computation graph. 
Signals remain binary throughout the network, enabling multiplication-free inference 
that maps directly to LUTs available on FPGAs. For the special case of $k=2$ (two binary
inputs per neuron), one recovers \emph{logic-gate networks} of \citet{petersen2022deep}. 

Although traditional weightless networks were hard to train due to their discrete, non-differentiable
structure, recently~\citet{bacellar2024differentiable} introduced a way to construct efficient 
surrogate gradients and a learnable interconnect, calling the resulting networks \emph{Differentiable Weightless Networks} (DWNs).
Below, we summarize the DWN forward and training mechanisms.

\myparagraph{Thermometer encoding.}
DWNs operate on binary signals, so any real-valued observation has to be discretized.
For any input $x\in\mathbb{R}$ and predefined thresholds 
$\tau_{1}<\cdots<\tau_{B}$, define the \emph{thermometer} encoding~\citep{carneiro2015multilingual},
\begin{equation}
E(x) = \big[{1}\{x \ge \tau_1\},\cdots,{1}\{x \ge \tau_{B}\}\big]\in\{0,1\}^{B}.
\end{equation}
For $d$-dimensional signals, each dimension is encoded individually, and the 
results are concatenated, yielding a $B \times d$-dimensional bitvector overall.

\myparagraph{Logic Layers.}
A DWN with $L$ layers comprises binary activation maps $b^{(\ell)}\in\{0,1\}^{D_\ell}$, $\ell=0,\dots,L$, where $D_{0}$ is the size of the encoded input, and for $\ell>0$, $D_\ell$ denotes the number of LUTs in layer $\ell$. Each LUT is a boolean function with arity $k$. For LUT $i$ we form an address $a^{(\ell+1)}_i\in\{0,1\}^k$ by selecting $k$ bits from $b^{(\ell)}$:%
\begin{equation}
a^{(\ell+1)}_i \;=\; \big(b^{(\ell)}_{c_1},\dots,b^{(\ell)}_{c_k}\big).
\end{equation}
The selection indices $(c_1,\dots,c_k)$ define the \emph{interconnect}. They are learned 
via straight-through estimation during training, as described in \citet{bacellar2024differentiable}. While \citet{bacellar2024differentiable} only train the first layer's interconnect, we also make the later ones learnable. This is inspired by results in \citet{kresse2025scalable}, where training accuracy consistently improves if later-layer interconnects are learnable.
Each LUT stores a table, $T$, of $2^k$ binary values. %
The output bit is the addressed entry,
\begin{equation}
b^{(\ell+1)}_i \;=\; T^{(\ell+1)}_i\big[\text{addr}(a^{(\ell+1)}_i)\big],
\end{equation}
where $\text{addr}(\cdot)$ maps the $k$-bit vector to its integer index. Concatenating all output bits yields $b^{(\ell+1)}$.

\myparagraph{Group aggregation.}
To allow for multi-dimensional outputs, a DWN's final binary 
features $b^{(L)}$ are partitioned into disjoint groups. 
For any group $G$, a \emph{group sum} is computed,
\(
s_G \;=\; \frac{1}{\tau}\sum_{i\in G} b^{(L)}_i,
\)
with a temperature $\tau>0$ used as a scale during training. 
For classification, these $s_G$ serve as logits for a softmax; 
at inference, the operation reduces to efficient popcount and 
argmax operations. 

\myparagraph{Gradient surrogate.}\label{sec:efd}
Because the forward pass is discrete, DWNs rely on surrogate gradients for training. 
We follow \citet{bacellar2024differentiable}, who employ an \emph{extended finite-difference} 
(EFD) estimator that aggregates contributions from all address locations. %

\section{Differentiable Weightless Controllers} %
\label{sec:dwn-rl}

\begin{figure}[t!]
\centering

\includegraphics{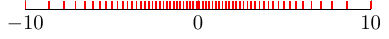}
\caption{\acronym thermometer threshold positions.}
\label{fig:thresholds}
\end{figure}

We now provide details on how \acronyms extend DWNs to continuous control tasks.
In terms of architecture, we adapt the input quantization, such that 
it can deal with changing input distributions during training, and 
we adapt the output layer to allow for multi-dimensional continuous 
actions.
Subsequently, we describe the RL training procedure. %

\subsection{Input Encoding}\label{sec:arch}

We map observations to binary inputs via clipped normalization and thermometer encoding. For each dimension $j$, we normalize with running mean and standard deviation, subsequently clipping: \(\hat{x}_j=\operatorname{clip}\big((x_j-\mu_j)/\sigma_j,\,-10,\,10\big)\). While in deep RL, normalization is often considered optional, we employ it \emph{always}, as we are projecting from $\mathbb{R}$ to a restricted, a priori known interval, for which we can subsequently choose thermometer thresholds $E_j$.  %
After normalization, we employ the same thermometer thresholds for all dimensions $d_{in}$.

For an \emph{odd} number of bits, $B$, we place thresholds at stretched-Gaussian quantiles: let $q_m=m/B$ for $m=1, \dots B-1$ and an additional quantile at \(\frac{1}{2}\), and set a stretch factor
\begin{equation}
s = \frac{10}{\big|\Phi^{-1}(\tfrac{1}{B})\big|},
\end{equation}
where \(\Phi^{-1} \) is the inverse cumulative probability distribution of the standard Gaussian. 
Define $\tau_{j,m}=s\,\Phi^{-1}(q_m)$ for $m=1,\dots,B$ so that the first/last thresholds land exactly at $\pm 10$, and the additional threshold at $0$. The thermometer code is $E_j(\hat{x}_j)=[{1}\{\hat{x}_j\ge\tau_{j,1}\},\ldots,{1}\{\hat{x}_j\ge\tau_{j,B}\}\,]\in\{0,1\}^{B}$, concatenated over $j$ to form $b^{(0)}$ for the first DWN layer. Figure \ref{fig:thresholds} illustrates the resulting input thresholds. %

\subsection{Continuous-control head}\label{sec:DWN-rl:ours}

To produce continuous actions we reinterpret the group aggregation as a bank of scalar heads, one per action dimension. Let $\{G_1,\dots,G_{d_{\text{act}}}\}$ be a partition of the final bits $b^{(L)}$. For dimension $d$ we first normalize the group sum: %
\begin{equation}
z_d \;=\; \frac{s_{G_d}}{|G_d|}\;-\;\frac{1}{2} \;\;\in[-\frac{1}{2},\frac{1}{2}].
\end{equation}
We then apply a per-dimension affine transformation:
\begin{equation}
l_d \;=\; \alpha_d\, z_d + \beta_d,
\end{equation}
with learnable scales $\alpha_d>0$ and bias $\beta_d\in\mathbb{R}$. 
To ensure positive $\alpha_d$, we parameterize $\alpha_d = e^{\alpha_{d,p}}$. 
This head is fully differentiable and integrates with policy-gradient objectives. 
The emitted $l_d$ is the logit, which is passed through an additional $\tanh$ 
in the case of SAC before computing the final action. 
The initialization value of $\alpha_d$ restricts the initial policy actions to 
a subinterval of the possible action space. This is comparable to initializing 
the final layer parameters with a low standard deviation, a strategy that has 
been shown to improve learning in RL~\citep{andrychowicz2020matters}.  

At deployment, the initial threshold operations required for the thermometer 
encoding should be implemented in a platform-dependent way.
Commonly, actual sensor readings are obtained as integers from an 
\emph{analog-to-digital converter} (ADC), in which the initial bitvector 
\(b^{(0)}\) can be computed as fixed integer-to-thermometer lookup per 
sensor channel (the constants used for normalization and clipping are 
fixed after training, so they can be folded into the threshold values).

Subsequently, actions can be computed using only LUT evaluations, popcounts 
and SRAM memory lookups:
starting with \(b^{(0)}\), the bitvectors propagate through \(L\) LUT layers 
to produce \(b^{(L)}\). 
For each action head \(d\), we popcount its group \(G_d\) to obtain the integer \(s_{G_d}\). 
An SRAM then implements the mapping from this popcount to the emitted control word, \ie, 
\(s_{G_d} \mapsto \alpha_d \bigl(s_{G_d}/|G_d| - \frac{1}{2}\bigr) + \beta_d\) and, for SAC, the subsequent \(\tanh\).
In practice, the table would output an integer actuator command. 

\subsection{Training}
The learnable components of \acronyms are the LUT entries, 
their connectivity, and the mapping from popcounts to 
action values. 
The latter are parameterized as an affine transformation, potentially 
followed by an additional hyperbolic tangent, so standard gradient-based 
learning is applicable. 
To learn the former two in a differentiable way, we rely on the connection learning and the EFD surrogate gradients 
of~\citet{bacellar2024differentiable}, making \acronyms overall compatible 
with any gradient-based reinforcement learning algorithm.

Note that, similar to previous work on quantized neural networks \citep{kresse2025learningquantizedcontinuouscontrollers}, during 
RL training we \emph{only} parametrize the policy networks as \acronyms,
because only these are required at deployment time. 
All auxiliary networks, such as the critic networks and the $\sigma_\theta$ 
head for SAC can remain as standard floating-point networks.

\section{Experiments}
\label{sec:experiments}
\begin{figure*}[t]
\centering

\includegraphics{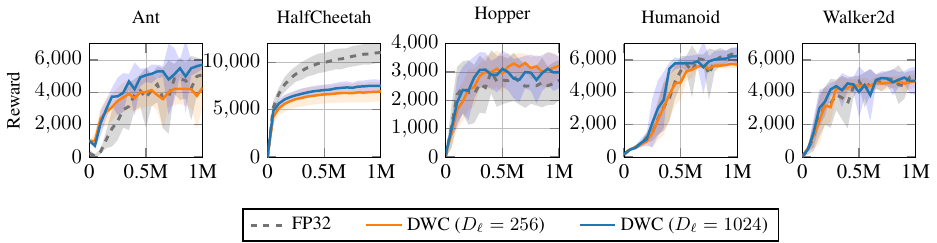}%

\caption{Mean return and standard deviation for ten models evaluated across the training steps, ten evaluation episodes per data point and model. Except for HalfCheetah (see main text), training trajectories are comparable to the FP baseline.}
\label{fig:training_steps}
\end{figure*}

\sisetup{
  detect-weight = true,
  detect-inline-weight = math,
  round-mode = places,
  round-precision = 1,
  scientific-notation = fixed,
  fixed-exponent = 3,
  exponent-to-prefix = true,
}

\newcommand{\numk}[1]{%
  \text{%
    \num[%
        detect-weight=true, 
        detect-inline-weight=text,
        mode=text,
        scientific-notation=false,
        round-mode=places,
        round-precision=1
    ]{\fpeval{#1/1000}}%
  }k%
}

\begin{table}[t]
\caption{Policy returns in different environments (median and the 25\% and 75\% quantiles over 10 trained models) 
for the proposed \acronyms, standard floating-point (FP), and low-precision quantized 
networks (Quant) of~\citet{kresse2025learningquantizedcontinuouscontrollers}. 
Highest values are marked in bold. 
\acronym returns are comparable to FP and at least as good as Quant across 
four of the five tasks (all except HalfCheetah).}
\label{tab:returns_lgn_vs_fp}
\centering
\small\addtolength{\tabcolsep}{-2pt}
\begin{tabular}{lllll}
\toprule
 \textbf{Environment\!\!\!\!\!} & \multicolumn{1}{c}{\textbf{FP}} &\multicolumn{1}{c}{\textbf{Quant}} & \multicolumn{1}{c}{\textbf{\acronym}} \\ %
\midrule
 Ant & \phantom{$1$}\numk{5598.251}$_{[\numk{4252.810}, \numk{5802.060}]}$ &  \phantom{$1$}\numk{4716.681}$_{[\numk{3887.572},\,\numk{4887.067}]}$ & {\bfseries\numk{5677.335}}$_{[\numk{5516.484},\,\numk{5905.774}]}$ \\
   HalfCheetah & {\bfseries\numk{11528.543}}$_{[\numk{10113.135}, \numk{11922.475}]}$ &  \numk{10465.013}$_{[\numk{9621.643},\,\numk{10955.758}]}$& \numk{7548.604}$_{[\numk{7096.618},\,\numk{7881.380}]}$ \\
   Hopper & \phantom{$1$}\numk{2796.813}$_{[\numk{2061.790}, \numk{3349.233}]}$ &   \phantom{$1$}\numk{1930.725}$_{[\numk{1096.022},\,\numk{3269.833}]}$ & {\bfseries\numk{3119.865}}$_{[\numk{2776.557},\,\numk{3386.147}]}$ \\
   Humanoid & {\bfseries\phantom{$1$}\numk{6185.556}}$_{[\numk{5956.489}, \numk{6650.238}]}$ &  \phantom{$1$}\numk{5954.057}$_{[\numk{5800.005},\,\numk{6054.452}]}$ & \numk{6140.962}$_{[\numk{5818.832},\,\numk{6605.264}]}$ \\
   Walker2d & {\bfseries\phantom{$1$}\numk{5043.828}}$_{[\numk{4697.417}, \numk{5194.093}]}$ &  \phantom{$1$}\numk{4656.477}$_{[\numk{4445.131},\,\numk{5019.122}]}$ & \numk{5024.956}$_{[\numk{4509.363},\,\numk{5196.075}]}$ \\

\bottomrule
\end{tabular}

\end{table}

\sisetup{
  scientific-notation = true,   %
  exponent-product   = \times,  %
  table-number-alignment = center
}

\renewcommand{\ns}[1]{%
\SI[scientific-notation=fixed, exponent-to-prefix=true, round-mode=places, round-precision=2]{#1}{} %
} %

\begin{table*}[t]
  \centering
  \begin{tabular}{
    l                         %
    l
    l                         %
    r                         %
    r                         %
    r                         %
    r                         %
    r       %
    l       %
    S[table-format=1.3e2]     %
    S[table-format=1.2e1]     %
  }
\toprule
    & \multicolumn{1}{c}{Environment} & \multicolumn{1}{c}{Reward} & \multicolumn{1}{c}{LUTs} & \multicolumn{1}{c}{FFs} & \multicolumn{1}{c}{B} & \multicolumn{1}{c}{DSP} & \multicolumn{1}{c}{Lat [\si{\micro\second}]} & \multicolumn{1}{c}{P [\si{\watt}]} & \multicolumn{1}{c}{TP} & \multicolumn{1}{c}{E.p.A. [\si{\joule}]}\\
    \midrule
    \multirow{5}{*}{\rotatebox{90}{\shortstack{$D_{\ell}=256$}}} & Ant & \phantom{$1$}\numk{4542.495}$_{[\numk{3425.491},\,\numk{5240.453}]}$& \numk{797} & \numk{530} & 0 & 0 & \ns{10}  & 0.105 & \num{100000000} & \num{1.05e-09} \\
    & HalfCheetah &\phantom{$1$}\numk{7054.791}$_{[\numk{6015.990},\,\numk{7859.055}]}$ &\numk{804} & \numk{354} & 0 & 0 & \ns{10}  & 0.105 & \num{100000000} & \num{1.05e-09} \\
    & Hopper & \phantom{$1$}{\bfseries\numk{3309.651}$_{[\numk{2964.868},\,\numk{3518.068}]}$}& \numk{902} & \numk{295} & 0 & 0 & \ns{10}  & 0.116 & \num{100000000} & \num{1.16e-09} \\
    & Humanoid &\phantom{$1$}\numk{5673.649}$_{[\numk{5544.689},\,\numk{5788.587}]}$& \numk{869} & \numk{1127} & 0 & 0 & \ns{10}  & 0.102 & \num{100000000} & \num{1.02e-09} \\
    & Walker2d &\phantom{$1$}\numk{4585.825}$_{[\numk{4526.129},\,\numk{4700.462}]}$& \numk{785} & \numk{392} & 0 & 0 & \ns{10}  & 0.107 & \num{100000000} & \num{1.07e-09} \\

\midrule
    \multirow{5}{*}{\rotatebox{90}{\shortstack{$D_{\ell}=1024$}}} & Ant &\phantom{$1$}{\bfseries\numk{5677.335}}$_{[\numk{5516.484},\,\numk{5905.774}]}$& \numk{3228} & \numk{1667} & 0 & 0 & \ns{20}  & 0.225 & \num{100000000} & \num{2.25e-09} \\
    & HalfCheetah &\phantom{$1$}\numk{7548.604}$_{[\numk{7096.618},\,\numk{7881.380}]}$& \numk{2954} & \numk{2150} & 0 & 0 & \ns{30}  & 0.208 & \num{100000000} & \num{2.08e-09} \\
    & Hopper & \phantom{$1$}{\numk{3119.865}}$_{[\numk{2776.557},\,\numk{3386.147}]}$ &\numk{3173} & \numk{2010} & 0 & 0 & \ns{30}  & 0.228 & \num{100000000} & \num{2.28e-09} \\
    & Humanoid & \phantom{$1$}{\bfseries\numk{6140.962}$_{[\numk{5818.832},\,\numk{6605.264}]}$} & \numk{3227} & \numk{3689} & 0 & 0 & \ns{20}  & 0.219 & \num{100000000} & \num{2.19e-09} \\
    & Walker2d & \phantom{$1$}{\bfseries\numk{5024.956}$_{[\numk{4509.363},\,\numk{5196.075}]}$} & \numk{2821} & \numk{2120} & 0 & 0 & \ns{30}  & 0.206 & \num{100000000} & \num{2.06e-09} \\

  \midrule\midrule
    \multirow{5}{*}{\rotatebox{90}{\shortstack{
 Kresse  \& \\
 Lampert \\
  \citeyearpar{kresse2025learningquantizedcontinuouscontrollers}
}}}
 & Ant         &  \phantom{$1$}\numk{4716.681}$_{[\numk{3887.572},\,\numk{4887.067}]}$ &\numk{2699} & \numk{4497} & 3   & 45 & \phantom{$00$}\ns{2290}   & 0.39 & \num{436681}   & \num{8.93e-7} \\
     & HalfCheetah &{\bfseries\numk{10465.013}$_{[\numk{9621.643},\,\numk{10955.758}]}$} & \numk{4336} & \numk{4565} & 15  & 11 & \ns{243230} & 0.33 & \num{4111}     & \num{8.02e-5} \\

     & Hopper      & \phantom{$1$}\numk{1930.725}$_{[\numk{1096.022},\,\numk{3269.833}]}$ &\numk{2377} & \numk{1994} & 0   & 45 & \phantom{$00$}\ns{210}    & 0.31 & \num{4761904}& \num{6.510001041600167e-8} \\
      & Humanoid   &\phantom{$1$}\numk{5954.057}$_{[\numk{5800.005},\,\numk{6054.452}]}$ & \numk{2256} & \numk{3118} & 1.5 & 45 & \phantom{$0$}\ns{15360}  & 0.33 & \num{65104}    & \num{5.068812976161219e-6} \\
     & Walker2d    &\phantom{$1$}\numk{4656.477}$_{[\numk{4445.131},\,\numk{5019.122}]}$ & \numk{1858} & \numk{1613} & 2   & 4  & \ns{162230} & 0.17 & \num{6164}     & \num{2.757949383517197e-5} \\
     
    \bottomrule
  \end{tabular}
  \caption{Post-synthesis resource utilization for one synthesized model, BRAM (B), end-to-end latency (Lat) in microseconds, estimated power (P) in Watts, peak throughput (TP) in actions per second, and energy per action (E.p.A.) in Joule on an Artix-7 {XC7A15T}\(-1\) at \(100\,\mathrm{MHz}\). Shown reward is the median over all ten models.}
  \label{tab:hardware}
\end{table*}

We evaluate the proposed \methods on five MuJoCo tasks~\cite{todorov2012mujoco}: Ant-v4, {HalfCheetah-v4}, {Hopper-v4}, Humanoid-v4 and {Walker2d-v4}, using SAC for training. Our code is publicly available.\footnote{\url{https://github.com/FKresse/differentiable_weightless_controllers}}
We also investigate returns for DDPG and PPO in Appendix~\ref{app:ddpgppo}. %

\myparagraph{Baselines.} As baselines, we use two weight-based neural network setups:
First, we use full-precision (FP) models trained with the {CleanRL} implementation~\cite{huang2022cleanrl} 
of SAC, as reported in \citet{kresse2025learningquantizedcontinuouscontrollers}. 
Here, the network has 256 neurons in the single hidden layer. 
In contrast to the original CleanRL implementation, the networks use running input normalization, as this has been found to improve return for SAC for our tasks. %
Second, we compare against the QAT-trained low-precision models (2- or 3-bit weights and activations) 
from \citet{kresse2025learningquantizedcontinuouscontrollers}. We use their smallest variants that achieve near-FP returns (e.g., Hopper: 16 hidden neurons).

\myparagraph{\acronym.} We adopt the same architectures as in the baseline implementations, except 
that we use \acronyms for the policy networks. Unless stated otherwise, \acronyms are instantiated 
with two layers of \(k{=}6\) input LUTs. 
We employ 1024 LUTs per layer, padding the last layer with LUTs to be divisible by the action dimension. 
Observations are discretized to 63 thermometer thresholds per dimension, as described in Section \ref{sec:dwn-rl}. 
Ablation studies can be found in Section~\ref{sec:ablation}. %

\myparagraph{Hyperparameters.} We adopt the protocol used in {CleanRL}; hence, we use the same hyperparameters across each algorithm for all investigated tasks.
For training the \acronyms we use the same hyperparameters as for the baselines, as specified in Appendix~\ref{app:hyperparams}.

\myparagraph{Training and Evaluation.} For each configuration, we train 10 models with different random seeds. Each model is trained for 1 million environment steps and then its undiscounted return is estimated from 1000 rollouts of the fixed policy from random starting states. 
Compared to the default {CleanRL} implementation, we perform all evaluation rollouts with the deterministic, maximum likelihood policy. We investigate the training overhead incured by \acronyms, as compared to floating-point networks, in Appendix \ref{app:training overhead}.

\subsection{Results: Quality}

\Cref{fig:training_steps} shows the training dynamics of \acronyms versus the FP baseline.
\Cref{tab:returns_lgn_vs_fp} shows the resulting returns after training, also including results for quantized networks.
Clearly, \acronyms learn policies of comparable quality to the floating-point networks, 
and hence they can readily serve as drop-in replacements, from a reward perspective. 
An exception is the HalfCheetah environment, where we observe a substantial return gap. 
Note that models with a cumulative reward of $7.5$k on HalfCheetah are not performing badly at all, 
indeed they learn to \emph{run}, \ie, master the environment. 
However, they do so at a slower pace than, \eg, the FP models with $11.5$k reward. 

Our findings confirm the observation from \citet{kresse2025learningquantizedcontinuouscontrollers}
that HalfCheetah is the task most resistant to network reduction and quantization, presumably 
because the task is \emph{capacity-limited}. We study this phenomenon further in Section~\ref{sec:ablation} and the Appendix. 

\subsection{Noise Robustness}
\begin{figure*}[t!]
    \centering

 \includegraphics{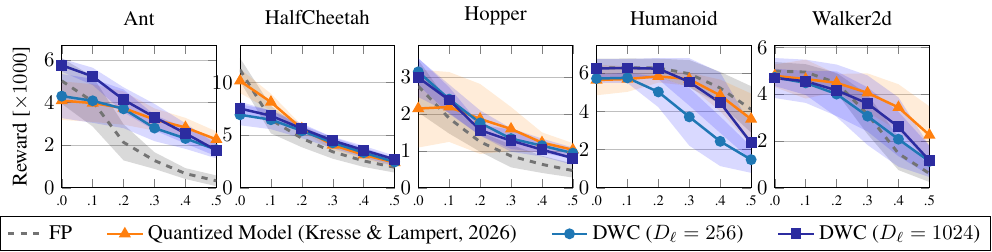}
 \caption{Reward performance under injected observation noise with varying noise level $\sigma$. Floating-point (FP), QAT policies from \citet{kresse2025learningquantizedcontinuouscontrollers} and our \acronyms on MuJoCo tasks. Bands show one standard deviation across trained models. The quantized models and \acronyms perform better than, or on par with, the FP baseline under injection, except for Humanoid, where the smaller \acronyms show reduced rewards for larger noise. }
 \label{fig:noise}
\end{figure*}
Following \citet{pmlr-v48-duan16}, we also assess the robustness of \acronyms 
to observation noise.
We inject zero-mean Gaussian noise with standard deviations \(\sigma\in\{0.1,\dots,0.5\}\) 
into normalized observations (unit variance) and compare \acronyms to the FP baseline and the quantized noise robustness results from \citet{kresse2025learningquantizedcontinuouscontrollers}. 
Figure~\ref{fig:noise} reports the results, which show that \acronyms achieve 
noise robustness profiles comparable to that of the quantized network,
and generally on the same level as, or better, than floating-point networks. The only exceptions are the compact ($D_{\ell}=256$) models, which perform worse on Humanoid under high noise. 

\subsection{Results: Efficiency}
The main advantage of \acronyms over standard deep neural network controllers is that they consist exclusively of elements that have direct representations on low-energy hardware platforms.
To demonstrate this, we perform synthesis and implementation (place-and-route) of the resulting networks for an FPGA, reporting required resources and energy estimates based on the manufacturer-provided tools.

We run out-of-context (OOC) synthesis and implementation with the Vivado toolchain (2022.2), targeting an Artix-7 XC7A15T (speed grade \(-1\)); see Table \ref{tab:artix} in Appendix \ref{app:fpga}
for the available device resources.
Note that the FPGA only has a total of \(10{,}400\) LUT-6s and twice this amount of flip-flops (FFs), 
which is a much smaller FPGA than in previous investigations of DWNs \citep{bacellar2024differentiable}. 
Instead, our setup is directly comparable to \citet{kresse2025learningquantizedcontinuouscontrollers}, 
who used the same reference FPGA and OOC synthesis. 

We target a clock frequency of \(100\,\si{\mega\hertz}\), inserting up to two pipeline stages---the 
first one between the two LUT layers, and the second one before the popcount---until we meet the 
desired timing.
As our design is implemented OOC, we do not consider any overhead due to I/O interfaces. 
Furthermore, in the main body of this work, we assume that all observation normalization steps and the final per-action lookups take place outside the synthesized core, because these depend on the specific application and interface
(\eg the input format depends on the type of ADCs, the output format on the type of DACs). Note that the quantized baseline \citep{kresse2025learningquantizedcontinuouscontrollers} likewise does not account for observation normalization. The final per-action mapping can be implemented as an additional single-cycle memory lookup (e.g., in BRAM/ROM), and would therefore add one cycle to the reported latency. An evaluation including input normalization and SRAM action lookup with representative interfaces is provided in Appendix \ref{sec:end-to-end}.

Table \ref{tab:hardware} reports reward, LUTs, FFs, BRAMs (B), latency (Lat), 
power (P), throughput (TP), and energy per action (E.p.A.) for each setup. 
Since full-precision policies are not practical on the selected hardware, 
we compare to the quantized networks from~\citet{kresse2025learningquantizedcontinuouscontrollers}.
The only difference to their setup is that our power estimates are 
based on the toolchain's post-implementation power report, rather than the 
less accurate Vivado Power Estimator.

The results show that \acronyms exhibit orders of magnitude lower latency and energy per action compared to the 
low-bitwidth quantized networks. %
For our standard $D_{\ell}=1024$ setup, latency is only 2 or 3 clock cycles, the throughput 
is the maximum possible at $10^8$ actions per second, and the energy usage is in
the range of 2 nanojoule per action.
In contrast, \citet{kresse2025learningquantizedcontinuouscontrollers} reports orders of 
magnitude higher and much more heterogeneous resource usage:
their latencies range from 21 cycles to over 24,000 cycles, their throughput between 
$4.1\times 10^3$ and $4.8\times 10^6$ actions per second, and their energy usage per action
between $2.8\times 10^{-5}$ and $6.5\times 10^{-8}$ Joule. Although control frequency is typically limited by actuation and sensing, a high control throughput can be useful, enabling multiple policy evaluations per step, e.g. for model-based lookahead or safety checks, when paired with a rollout model.

Additionally, in contrast to the quantized models, the computational core of \acronyms does not 
require any BRAM or DSP resources, making them deployable on even more limited hardware than 
the already small FPGA investigated here. 

To illustrate the scaling behavior, we include results for even smaller \acronyms  
with layer width $D_{\ell}=256$ in Table~\ref{tab:hardware}.
This results in single-clock-cycle latency, still maximum throughput, and energy
usage per action reduced further by a factor of approximately 2.

\section{Ablation Studies and Model Interpretability}\label{sec:ablation}
In this section we report ablation studies on the scaling behavior of
\acronyms with respect to their layer widths and the input size to the 
LUTs. Subsequently, we demonstrate how the sparse binary nature of \acronyms
allows for insights into their learned policies.

\subsection{Network Capacity}\label{sec:capacity}

We first investigate the effect of different network capacities. 
Concretely, we study the effect of varying the layer sizes (widths) and the number of inputs to each LUT. 
Further ablations on the impact of input resolution and number of layers are available in Appendix \ref{app:bits}.

\begin{figure*}[t!]
\centering
\includegraphics{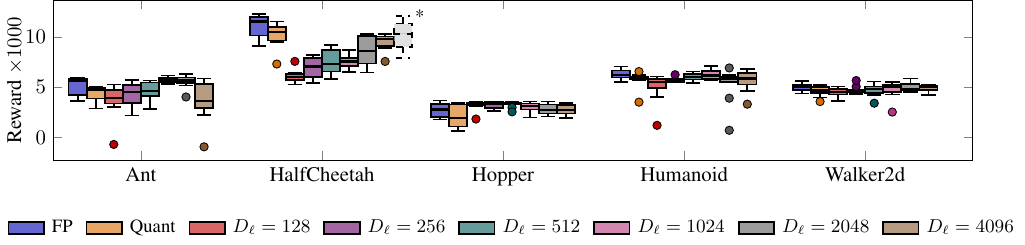}
\caption{Policy returns for FP, Quant and \acronyms with varying LUT layer widths. 
Generally, already models with 256 to 512 LUTs per layer achieve returns on par 
with the FP baseline. Only for HalfCheetah, we observe a monotonically increasing 
median return with increasing LUT layer width. * indicates a special high-capacity
model with 16k-LUTs per layer and 255-bit per input dimension, see Section~\ref{sec:ablation}.}
\label{fig:size ablation}
\end{figure*}

\myparagraph{Layer Width.}
Assuming that the number of layers is fixed, a large impact on the network capacity comes
from the layer width, $D_{\ell}$, \ie, the number of LUTs. 
We run experiments in the previous setting with layer width varying 
across \( \{128, 256, 512, 1024, 2048, 4096\} \), where the last 
layer is padded to be divisible by the action dimension,
and report the results in \Cref{fig:size ablation}.
For four of the tasks, the quality is relatively stable with 
respect to layer widths, with per-layer sizes above 256 generally 
exhibiting returns similar to the floating-point baseline.
This indicates that these learning tasks are not capacity limited, 
but that at the same time no overfitting effects seem to emerge.
As previously observed, HalfCheetah is an exception, where we 
observe that the quality of the policy increases monotonically with the width of the layers. 
There are two possible explanations for this phenomenon observed 
on HalfCheetah: 
(1) smoother actions, due to a higher output layer width, resulting in more possible actions, 
as each action head has a resolution of the size of the partition $G_d$; and 
(2) higher representation capacity and input layer resolution, as more LUTs can connect to different bits in the first layer. 
In light of the results in~\citet{kresse2025learningquantizedcontinuouscontrollers} that a 3-bit resolution suffices for HalfCheetah in the output actions; and hidden capacity 
appears to be the major bottleneck in quantized neural networks, we hypothesize that (2) 
is the case.

To explore this further, we investigate a substantially larger \acronym with $D_{\ell}=16,384$ LUTs 
per layer and inputs quantized to 255 levels (instead of 63). Due to the increased layer width, making the interconnect between the two LUT layers learnable requires $16,384^2\times k = 1.6$B parameters (with $k=6$), which is prohibitively memory-intensive. We therefore, only for this size, initialize this interconnect randomly and keep it fixed during training. 
Results are depicted in Figure~\ref{fig:size ablation} (dashed light gray entry).
With a median return value above $10.3k$, this setup now matches the returns of the quantized baseline (median return $10.4k$) 
and falls within the region of uncertainty of the floating-point policy (median return $11.5k$). 
Notably, even at the increased layer width, \acronyms require only $32$k \emph{lookups} and several popcounts--- %
substantially fewer than the over $70$k multiply-accumulate operations of the two baselines. %
We take this result as evidence that DWNs can scale to such tasks. %

\myparagraph{LUT Size.} 
Besides the number of LUTs, also the number of inputs per LUT, $k$, impacts the network capacity,
as more inputs imply that layers are more densely connected, and that the LUTs themselves
can express more complex relations. %
Figure~\ref{fig:k ablation} in the appendix reports average returns for all tasks with $k$ 
varied from \(\{2,3,4,5,6\}\) and constant layer width $D_{\ell}=1024$. In line with the previous experiments, 
we observe capacity effects limiting returns only for HalfCheetah, while the other returns remain 
largely unaffected.
This suggests that in practice, the number of LUT inputs can be chosen based on what matches the 
available hardware. 
For FPGAs, native support for $k=4$ or $k=6$ is common, whereas for custom 
ASICs, smaller values might be preferable~\cite{ahmed2000effect}.

\begin{figure*}[t!]
\centering
\includegraphics{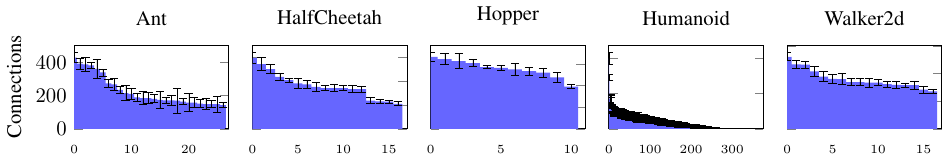}
\caption{Number of connections received for (sorted) input dimension averaged over trained \acronyms ($D_{\ell}=1024$) for all five environments.}
\label{fig:connections}
\end{figure*}

\begin{figure*}[t!]
\centering
\includegraphics{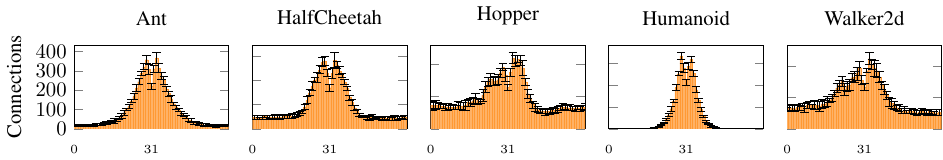}
\caption{Distribution of connections per threshold bit index across all environments. We consistently observe a two-modal distribution slightly left and right of the center (index 31, corresponding to normalized observation value 0).}
\label{fig:bitconnections}
\end{figure*}

\subsection{Diagnostics and Interpretability}
Deep networks are often criticized as \emph{black boxes}~\citep{molnar2020interpretable,vouros2022explainable}, 
where the trained policy offers little insight into which feature values drive specific actions.

Here, we demonstrate how binary sparse \acronyms can potentially contribute to overcoming this issue to some extent.
Because input bits correspond to specific input thresholds and connections to processing LUTs are both sparse and learned, we can infer feature importance simply by counting outgoing connections.
Figure~\ref{fig:connections} illustrates the number of connections received per observation dimension, averaged over 
the ten trained models ($D_{\ell}=1024$). The entries are sorted after aggregation according to the 
mean number of connections. 
The data shows that connectivity is non-uniform: some input dimensions receive substantially more connections 
than others, especially for higher-dimensional observation tasks, \ie Ant and Humanoid. 
The low standard deviations across independent models indicate that the network consistently identifies the same specific dimensions as relevant for solving the task.

Interestingly, for Humanoid, a large number of observation dimensions receive no 
connections at all (on average 275 out of 376 receive a connection). Because the 
trained model nevertheless attains performance comparable to the FP model, we conclude 
that the unconnected dimensions are not necessary for good control. %
In contrast, \emph{torso velocity observations} receive the highest number of 
connections for Humanoid. As the reward function is heavily dependent on forward velocity, this suggests the \acronym has identified the features most directly correlated with the reward signal. 

Figure~\ref{fig:bitconnections} shows the distribution of connections across 
input threshold bits, averaged over dimensions and over trained models. 
Recall that the input bit with index 31 out of 63 corresponds to the normalized 
observation value of 0, index 0 corresponds to the normalized value of $-10$ and index 63 to $10$ (Section~\ref{sec:dwn-rl}). 

Perhaps unsurprisingly, the highest connectivity density is typically found around 
the normalized observation value of zero. 
However, instead of a Normal distribution, we observe two distinct modes left and right of the center. This is potentially explained by the reduced probability density captured by the $0$ threshold, which is artificially inserted between the two thresholds to its left and right.
Furthermore, some tasks (Hopper, Walker, and HalfCheetah) show heavier tails than the others 
(Ant, Humanoid). 
We hypothesize that the heavier tails are explainable by the lower observation dimensionality 
of their respective tasks, and not due to an increased importance of extreme observation values. 
This hypothesis is supported by Figure~\ref{fig:bitconnections:128} in Appendix \ref{app:bits}, showing the 
same plot for \acronyms with $D_{\ell}=128$, which generally exhibit similar return performance (see Figure \ref{fig:size ablation}). 
Here, the tails are lighter, with the maximal number of possible first-layer connections having been significantly reduced to just $k\times D_{\ell}=768$, indicating that the heavier tails observed for the larger models are indeed due to the higher number of possible connections, suggesting that many connections might not be contributing significantly.

\section{Related Work}
\label{sec:related-work}

\myparagraph{Pruning and Quantization for RL.}  Previous work explored quantizing \citep{krishnan2019quarl} and pruning deep continuous control RL policies, or both simultaneously \citep{lu2024impact, ivanov2025neural}---which reduces their memory footprint and computational cost--- showing that substantial pruning (\(\geq 95\%\)) \citep{graesser2022state,tan2022rlx2} and quantization of most of the network to 3 or 2-bits is possible without harming policy returns \citep{kresse2025learningquantizedcontinuouscontrollers}. Additionally, there has been work on binary, one-bit, quantized RL policies \citep{valencia2019using, kadokawa2021binarized,chevtchenko2021combining, lazarus2022deepbinaryreinforcementlearning}. However, these focus on discrete action spaces or only partially binarize the network. In contrast, our \acronyms are fully one-bit policies for continuous control tasks. %

\myparagraph{Deep Boolean Networks (DBNs).} \citet{petersen2022deep} 
proposed \emph{Differentiable Logic Networks}, 
later extending them to a convolution‑style architecture~\cite{petersen2024convolutional}, 
training layers of 2-input LUTs (Boolean gates) end-to-end with gradient descent.

Since then, several extensions and variants of DBNs have been proposed to address different architectural bottlenecks. Most notably, the original formulation, which required $2^{2^k}$ parameters for $k$-input LUTs, has been adapted to only require $2^k$ parameters per LUT \citep{bacellar2024differentiable, ramirez2025llnn, gerlach2025warp}. Furthermore, interconnect learning was introduced \citep{bacellar2024differentiable,yue2024learninginterpretabledifferentiablelogic}, with extensions to improve training efficiency and similar constructions being proposed~\citep{kresse2025scalable, mommen2025method, fojcik2025lilogic}. Various other adaptations to improve scalability and convergence of DBNs have also been investigated \citep{kim2023stochastic, yousefi2025mindgapremovingdiscretization, kim2025towards, wang2026learning}.

Most commonly, DBNs have been applied to small-scale image and tabular classification tasks, with some preliminary work on discrete action reinforcement learning based on behavioral cloning~\citep{petersen2024rl}. Recently, this has been expanded to include anomaly detection \citep{gerlach2025rapid}, recurrent image generation \citep{miotti2025differentiable}, and recurrent language modeling \citep{buhrer2025recurrent}. In these applications, scalability has proven to be challenging, due to the high compute requirements during training, as in current work, at least four floating point parameters are required per limited expressivity, Boolean 2-input gate.

Finally, DBNs have shown to be promising for formal verification \citep{kresse2025}, an important consideration in control systems, which are often safety-critical. %

\myparagraph{Neuro-Symbolic Approaches for RL.} As our work touches upon the intersection of symbolic and neural methods, we very briefly review closely related work in this area. A more comprehensive overview can be found in \citep{acharya2023neurosymbolic}. \citet{bastani2018verifiable} investigate continuous control, distilling policies into decision trees, their approach scales at least to HalfCheetah. \citet{anderson2020neurosymbolic} project policies to a symbolic space. In both approaches, the policy is not directly learned with gradient descent as we do here. \citet{silva2020optimization} and \citet{kaptein2025interpretable} directly learn a decision tree; however, it is not clear if this approach scales to high-dimensional continuous tasks. In contrast to these works, \acronyms directly learn a Boolean structure with gradient descent, which scales to high-dimensional continuous control tasks such as Humanoid.

\section{Summary and Discussion}
\label{sec:disscusion}

In this work, we introduced \acronyms (\methods), which 
are differentiable weightless networks adapted to handle 
continuous input states and emit continuous actions. Furthermore, they can be trained using standard gradient-based 
reinforcement learning algorithms.
\acronyms allow for highly efficient implementation on 
low-energy embedded hardware, as we demonstrated by 
compiling them for an Artix-7 FPGA at 100 MHz, where the resulting
networks have extremely low resource requirements at substantially lower inference time 
(1--3 clock cycle latency) and nanojoule-level energy 
usage per action. 
At the same time, they can achieve return parity with 
standard floating-point networks even on high-dimensional, 
difficult RL tasks. %

We highlighted the interpretability and capacity properties of \acronyms 
through a sequence of experiments and ablation studies, particularly 
on the HalfCheetah environment, which is the most capacity-limited 
one of the studied MuJoCo tasks.

An existing limitation is the high computational cost at training time,
which significantly exceeds the cost at deployment, because of the 
relaxations required to allow for gradient-based training.
This aspect also means that training is currently only feasible 
in simulated environments, not interactively on-device.

\section*{Acknowledgments}

{This work was partially supported by the Austrian Science Fund (FWF) [10.55776/COE12] and the European Research Council (ERC-2020-AdG 101020093, VAMOS).} The research was supported by the Scientific Service Units (SSU) of ISTA through resources provided by Scientific Computing (SciComp). Finally, we thank the anonymous reviewers for their constructive feedback, which helped improve the evaluations presented in this manuscript.

\section*{Impact Statement}

This work advances machine learning for real-time control by enabling low-latency, energy-efficient policy inference on FPGA hardware. Potential impacts include reduced energy use and improved interpretability for edge deployment.

\bibliographystyle{icml2026}
\bibliography{references}

\begin{thebibliography}{59}
\providecommand{\natexlab}[1]{#1}
\providecommand{\url}[1]{\texttt{#1}}
\expandafter\ifx\csname urlstyle\endcsname\relax
  \providecommand{\doi}[1]{doi: #1}\else
  \providecommand{\doi}{doi: \begingroup \urlstyle{rm}\Url}\fi

\bibitem[Acharya et~al.(2023)Acharya, Raza, Dourado, Velasquez, and
  Song]{acharya2023neurosymbolic}
Acharya, K., Raza, W., Dourado, C., Velasquez, A., and Song, H.~H.
\newblock Neurosymbolic reinforcement learning and planning: A survey.
\newblock \emph{IEEE Transactions on Artificial Intelligence}, 2023.

\bibitem[Ahmed \& Rose(2004)Ahmed and Rose]{ahmed2000effect}
Ahmed, E. and Rose, J.
\newblock The effect of {LUT} and cluster size on deep-submicron {FPGA}
  performance and density.
\newblock \emph{IEEE Transactions on Very Large Scale Integration (VLSI)
  Systems}, 2004.

\bibitem[Akiba et~al.(2019)Akiba, Sano, Yanase, Ohta, and
  Koyama]{akiba2019optuna}
Akiba, T., Sano, S., Yanase, T., Ohta, T., and Koyama, M.
\newblock {O}ptuna: A next-generation hyperparameter optimization framework.
\newblock In \emph{ACM SIGKDD International Conference on Knowledge Discovery
  \& Data Mining}, 2019.

\bibitem[Aleksander et~al.(1984)Aleksander, Thomas, and
  Bowden]{aleksander1984wisard}
Aleksander, I., Thomas, W., and Bowden, P.
\newblock {WISARD}: A radical step forward in image recognition.
\newblock \emph{Sensor Review}, 1984.

\bibitem[Anderson et~al.(2020)Anderson, Verma, Dillig, and
  Chaudhuri]{anderson2020neurosymbolic}
Anderson, G., Verma, A., Dillig, I., and Chaudhuri, S.
\newblock Neurosymbolic reinforcement learning with formally verified
  exploration.
\newblock \emph{Conference on Neural Information Processing Systems (NeurIPS)},
  2020.

\bibitem[Andrychowicz et~al.(2021)Andrychowicz, Raichuk, Sta{\'n}czyk, Orsini,
  Girgin, Marinier, Hussenot, Geist, Pietquin, Michalski,
  et~al.]{andrychowicz2020matters}
Andrychowicz, M., Raichuk, A., Sta{\'n}czyk, P., Orsini, M., Girgin, S.,
  Marinier, R., Hussenot, L., Geist, M., Pietquin, O., Michalski, M., et~al.
\newblock What matters for on-policy deep actor-critic methods? a large-scale
  study.
\newblock In \emph{International Conference on Learning Representations
  (ICLR)}, 2021.

\bibitem[Bacellar et~al.(2024)Bacellar, Susskind, Breternitz~Jr, John, John,
  Lima, and Fran{\c{c}}a]{bacellar2024differentiable}
Bacellar, A.~T., Susskind, Z., Breternitz~Jr, M., John, E., John, L.~K., Lima,
  P.~M., and Fran{\c{c}}a, F.~M.
\newblock Differentiable weightless neural networks.
\newblock In \emph{International Conference on Machine Learning (ICML)}, 2024.

\bibitem[Bastani et~al.(2018)Bastani, Pu, and
  Solar-Lezama]{bastani2018verifiable}
Bastani, O., Pu, Y., and Solar-Lezama, A.
\newblock Verifiable reinforcement learning via policy extraction.
\newblock \emph{Conference on Neural Information Processing Systems (NeurIPS)},
  2018.

\bibitem[B{\"u}hrer et~al.(2025)B{\"u}hrer, Plesner, Aczel, and
  Wattenhofer]{buhrer2025recurrent}
B{\"u}hrer, S., Plesner, A., Aczel, T., and Wattenhofer, R.
\newblock Recurrent deep differentiable logic gate networks.
\newblock \emph{arXiv preprint arXiv:2508.06097}, 2025.

\bibitem[Carneiro et~al.(2015)Carneiro, Fran{\c{c}}a, and
  Lima]{carneiro2015multilingual}
Carneiro, H.~C., Fran{\c{c}}a, F.~M., and Lima, P.~M.
\newblock Multilingual part-of-speech tagging with weightless neural networks.
\newblock \emph{Neural Networks}, 2015.

\bibitem[Chevtchenko \& Ludermir(2021)Chevtchenko and
  Ludermir]{chevtchenko2021combining}
Chevtchenko, S.~F. and Ludermir, T.~B.
\newblock Combining {STDP} and binary networks for reinforcement learning from
  images and sparse rewards.
\newblock \emph{Neural Networks}, 2021.

\bibitem[Degrave et~al.(2022)Degrave, Felici, Buchli, Neunert, Tracey,
  Carpanese, Ewalds, Hafner, Abdolmaleki, de~las Casas, Donner, Fritz,
  Galperti, Huber, Keeling, Tsimpoukelli, Kay, Merle, Moret, Noury, Pesamosca,
  Pfau, Sauter, Sommariva, Coda, Duval, Fasoli, Kohli, Kavukcuoglu, Hassabis,
  and Riedmiller]{degrave2022magnetic}
Degrave, J., Felici, F., Buchli, J., Neunert, M., Tracey, B., Carpanese, F.,
  Ewalds, T., Hafner, R., Abdolmaleki, A., de~las Casas, D., Donner, C., Fritz,
  L., Galperti, C., Huber, A., Keeling, J., Tsimpoukelli, M., Kay, J., Merle,
  A., Moret, J.-M., Noury, S., Pesamosca, F., Pfau, D., Sauter, O., Sommariva,
  C., Coda, S., Duval, B., Fasoli, A., Kohli, P., Kavukcuoglu, K., Hassabis,
  D., and Riedmiller, M.
\newblock Magnetic control of tokamak plasmas through deep reinforcement
  learning.
\newblock \emph{Nature}, 2022.

\bibitem[Dosovitskiy et~al.(2021)Dosovitskiy, Beyer, Kolesnikov, Weissenborn,
  Zhai, Unterthiner, Dehghani, Minderer, Heigold, Gelly, Uszkoreit, and
  Houlsby]{dosovitskiy2020image}
Dosovitskiy, A., Beyer, L., Kolesnikov, A., Weissenborn, D., Zhai, X.,
  Unterthiner, T., Dehghani, M., Minderer, M., Heigold, G., Gelly, S.,
  Uszkoreit, J., and Houlsby, N.
\newblock An image is worth 16x16 words: Transformers for image recognition at
  scale.
\newblock In \emph{International Conference on Learning Representations
  (ICLR)}, 2021.

\bibitem[Duan et~al.(2016)Duan, Chen, Houthooft, Schulman, and
  Abbeel]{pmlr-v48-duan16}
Duan, Y., Chen, X., Houthooft, R., Schulman, J., and Abbeel, P.
\newblock Benchmarking deep reinforcement learning for continuous control.
\newblock In \emph{International Conference on Machine Learning (ICML)}, 2016.

\bibitem[Fojcik et~al.(2025)Fojcik, Zioma, and Armaitis]{fojcik2025lilogic}
Fojcik, K., Zioma, R., and Armaitis, J.
\newblock {LIL}ogic net: Compact logic gate networks with learnable
  connectivity for efficient hardware deployment.
\newblock \emph{arXiv preprint arXiv:2511.12340}, 2025.

\bibitem[Gerlach et~al.(2025{\natexlab{a}})Gerlach, Kauffman, V{\aa}ge, and
  Ojalvo]{gerlach2025rapid}
Gerlach, L., Kauffman, E., V{\aa}ge, L.~H., and Ojalvo, I.
\newblock Rapid inference of logic gate neural networks for anomaly detection
  in high energy physics.
\newblock \emph{arXiv preprint arXiv:2511.01908}, 2025{\natexlab{a}}.

\bibitem[Gerlach et~al.(2025{\natexlab{b}})Gerlach, V{\aa}ge, Gerlach, and
  Kauffman]{gerlach2025warp}
Gerlach, L., V{\aa}ge, L., Gerlach, T., and Kauffman, E.
\newblock {WARP}-{LUT}s:{W}alsh-assisted relaxation for probabilistic look up
  tables.
\newblock \emph{arXiv preprint arXiv:2510.15655}, 2025{\natexlab{b}}.

\bibitem[Gholami et~al.(2021)Gholami, Kim, Dong, Yao, Mahoney, and
  Keutzer]{gholami2022survey}
Gholami, A., Kim, S., Dong, Z., Yao, Z., Mahoney, M.~W., and Keutzer, K.
\newblock A survey of quantization methods for efficient neural network
  inference.
\newblock In \emph{Low-Power Computer Vision: Improve the Efficiency of
  Artificial Intelligence}. Chapman and Hall/CRC, 2021.

\bibitem[Graesser et~al.(2022)Graesser, Evci, Elsen, and
  Castro]{graesser2022state}
Graesser, L., Evci, U., Elsen, E., and Castro, P.~S.
\newblock The state of sparse training in deep reinforcement learning.
\newblock In \emph{International Conference on Machine Learning (ICML)}, 2022.

\bibitem[Haarnoja et~al.(2018)Haarnoja, Zhou, Abbeel, and
  Levine]{haarnoja2018soft}
Haarnoja, T., Zhou, A., Abbeel, P., and Levine, S.
\newblock Soft actor-critic: Off-policy maximum entropy deep reinforcement
  learning with a stochastic actor.
\newblock In \emph{International Conference on Machine Learning (ICML)}, 2018.

\bibitem[Huang et~al.(2022)Huang, Dossa, Ye, Braga, Chakraborty, Mehta, and
  Araújo]{huang2022cleanrl}
Huang, S., Dossa, R. F.~J., Ye, C., Braga, J., Chakraborty, D., Mehta, K., and
  Araújo, J.~G.
\newblock {CleanRL}: High-quality single-file implementations of deep
  reinforcement learning algorithms.
\newblock \emph{Journal of Machine Learning Research (JMLR)}, 2022.

\bibitem[{InvenSense Inc.}(2013)]{invensense2013mpu}
{InvenSense Inc.}
\newblock \emph{MPU-6000 and MPU-6050 Product Specification Revision 3.4},
  2013.

\bibitem[Ivanov et~al.(2025)Ivanov, Larionov, Maslennikov, and
  Voevodin]{ivanov2025neural}
Ivanov, D.~A., Larionov, D.~A., Maslennikov, O.~V., and Voevodin, V.~V.
\newblock Neural network compression for reinforcement learning tasks.
\newblock \emph{Scientific Reports}, 2025.

\bibitem[Kadokawa et~al.(2021)Kadokawa, Tsurumine, and
  Matsubara]{kadokawa2021binarized}
Kadokawa, Y., Tsurumine, Y., and Matsubara, T.
\newblock Binarized p-network: Deep reinforcement learning of robot control
  from raw images on {FPGA}.
\newblock \emph{IEEE Robotics and Automation Letters}, 2021.

\bibitem[Kaptein(2025)]{kaptein2025interpretable}
Kaptein, M.
\newblock Interpretable reinforcement learning for continuous action
  environments: Extending {DTPO} for continuous action spaces and evaluating
  competitiveness with {RPO}, 2025.

\bibitem[Kaufmann et~al.(2023)Kaufmann, Bauersfeld, Loquercio, M{\"u}ller,
  Koltun, and Scaramuzza]{kaufmann2023champion}
Kaufmann, E., Bauersfeld, L., Loquercio, A., M{\"u}ller, M., Koltun, V., and
  Scaramuzza, D.
\newblock Champion-level drone racing using deep reinforcement learning.
\newblock \emph{Nature}, 2023.

\bibitem[Kim(2023)]{kim2023stochastic}
Kim, Y.
\newblock Deep stochastic logic gate networks.
\newblock \emph{IEEE Access}, 2023.

\bibitem[Kim(2025)]{kim2025towards}
Kim, Y.
\newblock Towards narrowing the generalization gap in deep boolean networks.
\newblock \emph{Expert Systems with Applications}, 2025.

\bibitem[Kresse \& Lampert(2026)Kresse and
  Lampert]{kresse2025learningquantizedcontinuouscontrollers}
Kresse, F. and Lampert, C.~H.
\newblock Learning quantized continuous controllers for integer hardware.
\newblock In \emph{Learning for Dynamics \& Control Conference (L4DC)}, 2026.

\bibitem[Kresse et~al.(2025{\natexlab{a}})Kresse, Yu, and
  Lampert]{kresse2025scalable}
Kresse, F., Yu, E., and Lampert, C.~H.
\newblock Scalable interconnect learning in boolean networks.
\newblock \emph{arXiv preprint arXiv:2507.02585}, 2025{\natexlab{a}}.

\bibitem[Kresse et~al.(2025{\natexlab{b}})Kresse, Yu, Lampert, and
  Henzinger]{kresse2025}
Kresse, F., Yu, E., Lampert, C.~H., and Henzinger, T.~A.
\newblock Logic gate neural networks are good for verification.
\newblock In \emph{International Conference on Neuro-symbolic Systems (NeuS)},
  2025{\natexlab{b}}.

\bibitem[Krishnan et~al.(2022)Krishnan, Lam, Chitlangia, Wan, Barth-Maron,
  Faust, and Reddi]{krishnan2019quarl}
Krishnan, S., Lam, M., Chitlangia, S., Wan, Z., Barth-Maron, G., Faust, A., and
  Reddi, V.~J.
\newblock Qua{RL}: Quantization for fast and environmentally sustainable
  reinforcement learning.
\newblock \emph{Transactions on Machine Learning Research (TMLR)}, 2022.

\bibitem[Krizhevsky et~al.(2012)Krizhevsky, Sutskever, and
  Hinton]{krizhevsky2012imagenet}
Krizhevsky, A., Sutskever, I., and Hinton, G.~E.
\newblock Imagenet classification with deep convolutional neural networks.
\newblock In \emph{Conference on Neural Information Processing Systems
  (NeurIPS)}, 2012.

\bibitem[Lazarus \& Kochenderfer(2022)Lazarus and
  Kochenderfer]{lazarus2022deepbinaryreinforcementlearning}
Lazarus, C. and Kochenderfer, M.~J.
\newblock Deep binary reinforcement learning for scalable verification.
\newblock In \emph{International Conference on Intelligent Robots and Systems
  (IROS)}, 2022.

\bibitem[Lillicrap et~al.(2016)Lillicrap, Hunt, Pritzel, Heess, Erez, Tassa,
  Silver, and Wierstra]{lillicrap2015continuous}
Lillicrap, T.~P., Hunt, J.~J., Pritzel, A., Heess, N., Erez, T., Tassa, Y.,
  Silver, D., and Wierstra, D.
\newblock Continuous control with deep reinforcement learning.
\newblock In \emph{International Conference on Learning Representations
  (ICLR)}, 2016.

\bibitem[Lu et~al.(2024)Lu, Alemi, and Rawassizadeh]{lu2024impact}
Lu, H., Alemi, M., and Rawassizadeh, R.
\newblock The impact of quantization and pruning on deep reinforcement learning
  models.
\newblock \emph{arXiv preprint arXiv:2407.04803}, 2024.

\bibitem[Ludermir \& {de Oliveira}(1994)Ludermir and {de
  Oliveira}]{ludermir1994weightless}
Ludermir, T.~B. and {de Oliveira}, W.~R.
\newblock Weightless neural models.
\newblock \emph{Computer Standards \& Interfaces}, 1994.

\bibitem[Miller et~al.(2025)Miller, Yu, Brauckmann, and
  Farshidian]{miller2025high}
Miller, A., Yu, F., Brauckmann, M., and Farshidian, F.
\newblock High-performance reinforcement learning on spot: Optimizing
  simulation parameters with distributional measures.
\newblock In \emph{ICRA}, 2025.

\bibitem[Miotti et~al.(2025)Miotti, Niklasson, Randazzo, and
  Mordvintsev]{miotti2025differentiable}
Miotti, P., Niklasson, E., Randazzo, E., and Mordvintsev, A.
\newblock Differentiable logic cellular automata: From game of life to pattern
  generation.
\newblock In \emph{Conference on Artificial Life (ALIFE)}, 2025.

\bibitem[Mnih et~al.(2013)Mnih, Kavukcuoglu, Silver, Graves, Antonoglou,
  Wierstra, and Riedmiller]{mnih2013playing}
Mnih, V., Kavukcuoglu, K., Silver, D., Graves, A., Antonoglou, I., Wierstra,
  D., and Riedmiller, M.
\newblock Playing {Atari} with deep reinforcement learning.
\newblock \emph{arXiv preprint arXiv:1312.5602}, 2013.

\bibitem[Molnar(2020)]{molnar2020interpretable}
Molnar, C.
\newblock \emph{Interpretable machine learning}.
\newblock lulu.com, 2020.

\bibitem[Mommen et~al.(2025)Mommen, Keuninckx, Hartmann, and
  Wambacq]{mommen2025method}
Mommen, W., Keuninckx, L., Hartmann, M., and Wambacq, P.
\newblock A method for optimizing connections in differentiable logic gate
  networks.
\newblock \emph{arXiv preprint arXiv:2507.06173}, 2025.

\bibitem[Petersen et~al.(2022)Petersen, Borgelt, Kuehne, and
  Deussen]{petersen2022deep}
Petersen, F., Borgelt, C., Kuehne, H., and Deussen, O.
\newblock Deep differentiable logic gate networks.
\newblock In \emph{Conference on Neural Information Processing Systems
  (NeurIPS)}, 2022.

\bibitem[Petersen et~al.(2024{\natexlab{a}})Petersen, Borgelt, and
  Ermon]{petersen2024rl}
Petersen, F., Borgelt, C., and Ermon, S.
\newblock Efficient reinforcement learning agents with differentiable logic
  gate networks.
\newblock In \emph{Differentiable Optimization Everywhere: Simulation,
  Estimation, Learning, and Control (DiffOpt), Workshop at Conference on Robot
  Learning (CoRL)}, 2024{\natexlab{a}}.

\bibitem[Petersen et~al.(2024{\natexlab{b}})Petersen, Kuehne, Borgelt, Welzel,
  and Ermon]{petersen2024convolutional}
Petersen, F., Kuehne, H., Borgelt, C., Welzel, J., and Ermon, S.
\newblock Convolutional differentiable logic gate networks.
\newblock In \emph{Conference on Neural Information Processing Systems
  (NeurIPS)}, 2024{\natexlab{b}}.

\bibitem[Radford et~al.(2019)Radford, Wu, Child, Luan, Amodei, and
  Sutskever]{radford2019language}
Radford, A., Wu, J., Child, R., Luan, D., Amodei, D., and Sutskever, I.
\newblock Language models are unsupervised multitask learners.
\newblock \emph{OpenAI blog}, 2019.

\bibitem[Ram{\'\i}rez et~al.(2025)Ram{\'\i}rez, Garcia-Espinosa, Concha,
  Aranda, and Schiavi]{ramirez2025llnn}
Ram{\'\i}rez, I., Garcia-Espinosa, F.~J., Concha, D., Aranda, L.~A., and
  Schiavi, E.
\newblock {LLNN}: A scalable {LUT}-based logic neural network architecture for
  {FPGA}s.
\newblock \emph{IEEE Transactions on Circuits and Systems I: Regular Papers
  (TCAS-I)}, 2025.

\bibitem[Schulman et~al.(2017)Schulman, Wolski, Dhariwal, Radford, and
  Klimov]{schulman2017proximal}
Schulman, J., Wolski, F., Dhariwal, P., Radford, A., and Klimov, O.
\newblock Proximal policy optimization algorithms.
\newblock \emph{arXiv preprint arXiv:1707.06347}, 2017.

\bibitem[Silva et~al.(2020)Silva, Killian, Jimenez, Son, and
  Gombolay]{silva2020optimization}
Silva, A., Killian, T., Jimenez, I., Son, S.-H., and Gombolay, M.
\newblock Optimization methods for interpretable differentiable decision trees
  applied to reinforcement learning.
\newblock In \emph{International Conference on Artificial Intelligence and
  Statistics (AISTATS)}, 2020.

\bibitem[Silver et~al.(2016)Silver, Huang, Maddison, Guez, Sifre, van~den
  Driessche, Schrittwieser, Antonoglou, Panneershelvam, Lanctot, Dieleman,
  Grewe, Nham, Kalchbrenner, Sutskever, Lillicrap, Leach, Kavukcuoglu, Graepel,
  and Hassabis]{silver2016mastering}
Silver, D., Huang, A., Maddison, C.~J., Guez, A., Sifre, L., van~den Driessche,
  G., Schrittwieser, J., Antonoglou, I., Panneershelvam, V., Lanctot, M.,
  Dieleman, S., Grewe, D., Nham, J., Kalchbrenner, N., Sutskever, I.,
  Lillicrap, T.~P., Leach, M., Kavukcuoglu, K., Graepel, T., and Hassabis, D.
\newblock Mastering the game of {G}o with deep neural networks and tree search.
\newblock \emph{Nature}, 2016.

\bibitem[Sutton \& Barto(2018)Sutton and Barto]{barto2021reinforcement}
Sutton, R.~S. and Barto, A.~G.
\newblock \emph{Reinforcement Learning: An Introduction}.
\newblock MIT Press, 2018.

\bibitem[Tan et~al.(2023)Tan, Hu, Pan, Huang, and Huang]{tan2022rlx2}
Tan, Y., Hu, P., Pan, L., Huang, J., and Huang, L.
\newblock {RL}x2: Training a sparse deep reinforcement learning model from
  scratch.
\newblock In \emph{International Conference on Learning Representations
  (ICLR)}, 2023.

\bibitem[Todorov et~al.(2012)Todorov, Erez, and Tassa]{todorov2012mujoco}
Todorov, E., Erez, T., and Tassa, Y.
\newblock {MuJoCo}: A physics engine for model-based control.
\newblock In \emph{International Conference on Intelligent Robots and Systems
  (IROS)}, 2012.

\bibitem[Valencia et~al.(2019)Valencia, Sham, and Sinnen]{valencia2019using}
Valencia, R., Sham, C.-W., and Sinnen, O.
\newblock Using neuroevolved binary neural networks to solve reinforcement
  learning environments.
\newblock In \emph{IEEE Asia Pacific Conference on Circuits and Systems
  (APCCAS)}, 2019.

\bibitem[Vaswani et~al.(2017)Vaswani, Shazeer, Parmar, Uszkoreit, Jones, Gomez,
  Kaiser, and Polosukhin]{vaswani2017attention}
Vaswani, A., Shazeer, N., Parmar, N., Uszkoreit, J., Jones, L., Gomez, A.~N.,
  Kaiser, L., and Polosukhin, I.
\newblock Attention is all you need.
\newblock In \emph{Conference on Neural Information Processing Systems
  (NeurIPS)}, 2017.

\bibitem[Vouros(2022)]{vouros2022explainable}
Vouros, G.~A.
\newblock Explainable deep reinforcement learning: state of the art and
  challenges.
\newblock \emph{ACM Computing Surveys}, 2022.

\bibitem[Wang et~al.(2026)Wang, Mao, Zhang, and Vechev]{wang2026learning}
Wang, S., Mao, Y., Zhang, Y., and Vechev, M.
\newblock Learning compact boolean networks.
\newblock \emph{arXiv preprint arXiv:2602.05830}, 2026.

\bibitem[Yousefi et~al.(2025)Yousefi, Plesner, Aczel, and
  Wattenhofer]{yousefi2025mindgapremovingdiscretization}
Yousefi, S., Plesner, A., Aczel, T., and Wattenhofer, R.
\newblock Mind the gap: Removing the discretization gap in differentiable logic
  gate networks.
\newblock In \emph{Conference on Neural Information Processing Systems
  (NeurIPS)}, 2025.

\bibitem[Yue \& Jha(2024)Yue and
  Jha]{yue2024learninginterpretabledifferentiablelogic}
Yue, C. and Jha, N.~K.
\newblock Learning interpretable differentiable logic networks.
\newblock \emph{{IEEE} Transactions on Circuits and Systems for Artificial
  Intelligence}, 2024.

\end{thebibliography}

\appendix
\onecolumn

\section{DDPG and PPO returns}

\begin{table*}[t!]
\centering
\small
\begin{tabular}{l l l l l l}
\toprule
\textbf{Alg.} & \textbf{Environment} & \textbf{FP $\pm$ sd} & \textbf{\acronym{} $\pm$ sd} & \textbf{FP median [q\textsubscript{25}, q\textsubscript{75}]} & \textbf{\acronym{} median [q\textsubscript{25}, q\textsubscript{75}]} \\
\midrule
\multirow{5}{*}{\rotatebox{90}{\textbf{SAC}}} & Ant-v4 & \numk{5089.344}$_{\pm\,\numk{943.045}}$ & \numk{5707.631}$_{\pm\,\numk{272.326}}$ & \numk{5598.251}$_{[\numk{4252.810}, \numk{5802.060}]}$ & \numk{5677.335}$_{[\numk{5516.484},\,\numk{5905.774}]}$ \\
  & HalfCheetah-v4 & \numk{11066.069}$_{\pm\,\numk{1190.071}}$ & \numk{7528.745}$_{\pm\,\numk{678.601}}$ & \numk{11528.543}$_{[\numk{10113.135}, \numk{11922.475}]}$ & \numk{7548.604}$_{[\numk{7096.618},\,\numk{7881.380}]}$ \\
  & Hopper-v4 & \numk{2710.482}$_{\pm\,\numk{711.618}}$ & \numk{2989.452}$_{\pm\,\numk{553.400}}$ & \numk{2796.813}$_{[\numk{2061.790}, \numk{3349.233}]}$ & \numk{3119.865}$_{[\numk{2776.557},\,\numk{3386.147}]}$ \\
  & Humanoid-v4 & \numk{6261.652}$_{\pm\,\numk{512.506}}$ & \numk{6229.831}$_{\pm\,\numk{505.760}}$ & \numk{6185.556}$_{[\numk{5956.489}, \numk{6650.238}]}$ & \numk{6140.962}$_{[\numk{5818.832},\,\numk{6605.264}]}$ \\
  & Walker2d-v4 & \numk{4976.808}$_{\pm\,\numk{395.305}}$ & \numk{4707.204}$_{\pm\,\numk{861.232}}$ & \numk{5043.828}$_{[\numk{4697.417}, \numk{5194.093}]}$ & \numk{5024.956}$_{[\numk{4509.363},\,\numk{5196.075}]}$ \\
\midrule
\multirow{5}{*}{\rotatebox{90}{\textbf{DDPG}}} & Ant-v4 & \numk{1068.462}$_{\pm\,\numk{608.926}}$ & \numk{1317.209}$_{\pm\,\numk{905.886}}$ & \numk{962.309}$_{[\numk{551.646}, \numk{1449.853}]}$ & \numk{1408.813}$_{[\numk{1030.861},\,\numk{1641.696}]}$ \\
  & HalfCheetah-v4 & \numk{11156.418}$_{\pm\,\numk{637.159}}$ & \numk{7372.269}$_{\pm\,\numk{796.289}}$ & \numk{11278.164}$_{[\numk{10907.899}, \numk{11433.200}]}$ & \numk{7293.430}$_{[\numk{6741.578},\,\numk{7911.999}]}$ \\
  & Hopper-v4 & \numk{2300.168}$_{\pm\,\numk{799.100}}$ & \numk{1957.589}$_{\pm\,\numk{355.643}}$ & \numk{2170.331}$_{[\numk{1975.914}, \numk{2962.614}]}$ & \numk{1950.127}$_{[\numk{1824.361},\,\numk{2183.376}]}$ \\
  & Humanoid-v4 & \numk{1927.305}$_{\pm\,\numk{600.498}}$ & \numk{1253.621}$_{\pm\,\numk{229.313}}$ & \numk{1654.671}$_{[\numk{1460.821}, \numk{2409.503}]}$ & \numk{1187.186}$_{[\numk{1147.357},\,\numk{1243.349}]}$ \\
  & Walker2d-v4 & \numk{1718.485}$_{\pm\,\numk{530.599}}$ & \numk{2082.731}$_{\pm\,\numk{781.582}}$ & \numk{1544.185}$_{[\numk{1435.290}, \numk{1762.262}]}$ & \numk{2327.243}$_{[\numk{1393.659},\,\numk{2615.407}]}$ \\
\midrule
\multirow{5}{*}{\rotatebox{90}{\textbf{PPO}}} & Ant-v4 & \numk{879.215}$_{\pm\,\numk{390.197}}$ & \numk{1646.641}$_{\pm\,\numk{148.708}}$ & \numk{710.888}$_{[\numk{641.761}, \numk{909.925}]}$ & \numk{1632.095}$_{[\numk{1522.825},\,\numk{1734.212}]}$ \\
  & HalfCheetah-v4 & \numk{1636.110}$_{\pm\,\numk{573.374}}$ & \numk{2073.549}$_{\pm\,\numk{593.281}}$ & \numk{1492.209}$_{[\numk{1373.433}, \numk{1603.063}]}$ & \numk{2141.076}$_{[\numk{1621.672},\,\numk{2499.845}]}$ \\
  & Hopper-v4 & \numk{1909.266}$_{\pm\,\numk{579.284}}$ & \numk{1673.763}$_{\pm\,\numk{571.398}}$ & \numk{2078.171}$_{[\numk{1522.170}, \numk{2367.507}]}$ & \numk{1868.189}$_{[\numk{1541.255},\,\numk{2036.373}]}$ \\
  & Humanoid-v4 & \numk{533.846}$_{\pm\,\numk{44.968}}$ & \numk{496.740}$_{\pm\,\numk{30.451}}$ & \numk{540.217}$_{[\numk{488.500}, \numk{562.626}]}$ & \numk{502.235}$_{[\numk{477.777},\,\numk{509.293}]}$ \\
  & Walker2d-v4 & \numk{2177.542}$_{\pm\,\numk{1066.378}}$ & \numk{1374.326}$_{\pm\,\numk{337.495}}$ & \numk{2395.381}$_{[\numk{1288.875}, \numk{2890.953}]}$ & \numk{1307.243}$_{[\numk{1075.360},\,\numk{1582.798}]}$ \\
\bottomrule
\end{tabular}
\caption{Return performance for \acronyms and FP baseline. Showing mean $\pm$ standard deviation and median [25\textsuperscript{th} percentile, 75\textsuperscript{th} percentile] over 10 trained models. In contrast to SAC, results for DDPG and PPO are more inconsistent, with \acronyms sometimes exceeding and sometimes falling short of FP performance.}
\label{tab:fp-vs-dwn-stats}
\end{table*}

\label{app:ddpgppo}

In this appendix we provide results for training \acronyms with DDPG and PPO. Table \ref{tab:fp-vs-dwn-stats} shows results for all investigated tasks and algorithms. For SAC and DDPG, we use the same hyperparameters for both \acronyms and the FP32 baseline. \acronyms trained with DDPG use 1024 LUTs per layer. For PPO we perform hyperparameter tuning for both FP32 and \acronyms separately, resulting in 256 LUTs per layer for \acronyms. Hyperparameters and search details are provided in Appendix \ref{app:hyperparams}.

For PPO, we directly use the generated logit $l_a$ as the action, whereas for DDPG, we compute the action by applying a $\tanh$ as in SAC.

For the DDPG FP32 baseline, we report results with \emph{unnormalized} observations, as this outperforms observation normalization on our tasks~\citep{kresse2025learningquantizedcontinuouscontrollers}. All other experiments (SAC, PPO) use observation normalization also for the FP model.
While SAC shows very consistent results for FP and \acronyms across all tasks, DDPG and PPO results vary, with \acronyms sometimes outperforming and sometimes underperforming FP32. Overall, for DDPG the performance gap for HalfCheetah remains the largest, similar to SAC. 

\section{End-to-end synthesis}
\label{sec:end-to-end}

\sisetup{
  scientific-notation = true,   %
  exponent-product    = \times,  %
  table-number-alignment = center
}

\renewcommand{\ns}[1]{%
\SI[scientific-notation=fixed, exponent-to-prefix=true, round-mode=places, round-precision=2]{#1}{} %
} %

\begin{table*}[t]
  \centering
  \begin{tabular}{
    l                         %
    l
    l                         %
    r                         %
    r                         %
    r                         %
    r                         %
    r       %
    l       %
    S[table-format=1.3e2]     %
    S[table-format=1.2e1]     %
  }
\toprule
    & \multicolumn{1}{c}{Environment} & \multicolumn{1}{c}{Reward} & \multicolumn{1}{c}{LUTs} & \multicolumn{1}{c}{FFs} & \multicolumn{1}{c}{B} & \multicolumn{1}{c}{DSP} & \multicolumn{1}{c}{Lat [\si{\micro\second}]} & \multicolumn{1}{c}{P [\si{\watt}]} & \multicolumn{1}{c}{TP} & \multicolumn{1}{c}{E.p.A. [\si{\joule}]}\\
    \midrule
    \multirow{5}{*}{\rotatebox{90}{\shortstack{$b_{\text{obs}}=\text{12-bit}$}}} & Ant & \phantom{$1$}\numk{4200}$_{[\numk{3500},\,\numk{5200}]}$& \numk{3528} & \numk{616} & 4 & 0 & \ns{30}  & 0.131 & \num{100000000} & \num{1.31e-09} \\
    & HalfCheetah &\phantom{$1$}\numk{6900}$_{[\numk{6000},\,\numk{7900}]}$ &\numk{2582} & \numk{479} & 3 & 0 & \ns{30}  & 0.125 & \num{100000000} & \num{1.25e-09} \\
    & Hopper & \phantom{$1$}\numk{3200}$_{[\numk{2900},\,\numk{3600}]}$& \numk{2344} & \numk{381} & 1.5 & 0 & \ns{0.03}  & 0.124 & \num{100000000} & \num{1.24e-09} \\
    & Humanoid &\phantom{$1$}\numk{5700}$_{[\numk{5600},\,\numk{5800}]}$& \numk{6457} & \numk{2724} & 8.5 & 0 & \ns{20}  & 0.164 & \num{100000000} & \num{1.64e-09} \\
    & Walker2d &\phantom{$1$}\numk{4700}$_{[\numk{4500},\,\numk{4700}]}$& \numk{2784} & \numk{460} & 3 & 0 & \ns{0.03}  & 0.125 & \num{100000000} & \num{1.25e-09} \\

\midrule
    \multirow{5}{*}{\rotatebox{90}{\shortstack{$b_{\text{obs}}=\text{16-bit}$}}} & Ant &\phantom{$1$}\numk{4300}$_{[\numk{3500},\,\numk{5300}]}$& \numk{4470} & \numk{724} & 4 & 0 & \ns{30}  & 0.137 & \num{100000000} & \num{1.37e-09} \\
    & HalfCheetah &\phantom{$1$}\numk{6900}$_{[\numk{6000},\,\numk{7800}]}$& \numk{3219} & \numk{547} & 3 & 0 & \ns{30}  & 0.130 & \num{100000000} & \num{1.30e-09} \\
    & Hopper & \phantom{$1$}\numk{3200}$_{[\numk{3000},\,\numk{3600}]}$ &\numk{2891} & \numk{425} & 1.5 & 0 & \ns{30}  & 0.129 & \num{100000000} & \num{1.29e-09} \\
    & Humanoid & \phantom{$1$}\numk{5700}$_{[\numk{5500},\,\numk{5800}]}$ & \numk{8466} & \numk{3619} & 8.5 & 0 & \ns{20}  & 0.181 & \num{100000000} & \num{1.81e-09} \\
    & Walker2d & \phantom{$1$}\numk{4700}$_{[\numk{4400},\,\numk{4600}]}$ & \numk{3484} & \numk{528} & 3 & 0 & \ns{30}  & 0.130 & \num{100000000} & \num{1.30e-09} \\
     
    \bottomrule
  \end{tabular}
  \caption{Post-synthesis resource utilization for 12-bit and 16-bit signed observations, showing BRAM (B), end-to-end latency (Lat) in microseconds, estimated Power (P) in Watts, peak throughput (TP) in actions per second, and energy per action (E.p.A.) in Joule on an Artix-7 {XC7A15T}\(-1\) at \(100\,\mathrm{MHz}\). All models have two layers with $D_\ell=256$. }
  \label{tab:end-to-end}
\end{table*}

\sisetup{
  scientific-notation = true,   %
  exponent-product    = \times,  %
  table-number-alignment = center
}

\renewcommand{\ns}[1]{%
\SI[scientific-notation=fixed, exponent-to-prefix=true, round-mode=places, round-precision=2]{#1}{} %
} %

\begin{table*}[t]
  \centering
  \begin{tabular}{
    l                         %
    l
    r                         %
    r                         %
    r                         %
    r                         %
    r       %
    l       %
    S[table-format=1.3e2]     %
    S[table-format=1.2e1]     %
  }
\toprule
    & \multicolumn{1}{c}{Environment} & \multicolumn{1}{c}{LUTs} & \multicolumn{1}{c}{FFs} & \multicolumn{1}{c}{B} & \multicolumn{1}{c}{DSP} & \multicolumn{1}{c}{Lat [\si{\micro\second}]} & \multicolumn{1}{c}{P [\si{\watt}]} & \multicolumn{1}{c}{TP} & \multicolumn{1}{c}{E.p.A. [\si{\joule}]}\\
    \midrule
    \multirow{3}{*}{\rotatebox{90}{\shortstack{FP}}} 
    & HalfCheetah & \numk{7293} & \numk{8973} & 3   & 34 & \ns{200} & 0.259 & \num{100000000} & \num{2.59e-09} \\
    & Hopper      & \numk{4847} & \numk{5518} & 1.5 & 22 & \ns{200} & 0.191 & \num{100000000} & \num{1.91e-09} \\
    & Walker2d    & \numk{7588} & \numk{8957} & 3   & 34 & \ns{200} & 0.261 & \num{100000000} & \num{2.61e-09} \\
    \bottomrule
  \end{tabular}
  \caption{Post-synthesis resource utilization, assuming the unusual setup where quantization to signed integers is performed on-device, and the FPGA is provided with FP32 values. Showing BRAM (B), end-to-end latency (Lat) in microseconds, estimated Power (P) in Watts, peak throughput (TP) in actions per second, and energy per action (E.p.A.) in Joule on an Artix-7 {XC7A15T}\(-1\) at \(100\,\mathrm{MHz}\). Two layers with $D_\ell=256$.}
  \label{tab:fp-included}
\end{table*}

Here, we provide synthesis results that account for the observation normalization and final affine mapping. Results for $D_\ell = 256$ and two representative quantization bitwidths of the sensors (observations) are given in Table \ref{tab:end-to-end}. The place-and-route is still performed OOC, and all sensor values are assumed to be already present on the FPGA. We assume sensor bit-widths $b_{\text{obs}}=\{12,16\}$, representing typical values; for instance, see the MPU-6000 IMU commonly used in UAVs~\citep{invensense2013mpu}.

As all observation values in the MuJoCo simulator are provided as floating-point values, we chose to perform post-training symmetric quantization of the observation values (without a zero-point). Hence, for $b_{\text{obs}}$ bits, 
\begin{equation}
    Q_{\text{max}}= 2^{b_{\text{obs}}-1}-1  \qquad Q_S = \frac{x_{\text{max}}}{Q_{\text{max}}}\qquad Q(x) =  \text{clip}\bigg(\bigg\lfloor \frac{x}{Q_{S}}\bigg\rceil, -Q_{\text{max}}, Q_{\text{max}}\bigg),
\end{equation}

with $Q_{\text{max}}$ being the largest integer representable in the quantized domain, $Q_S$ the quantization scale, and $Q(x)$ the quantization operation. The term $x_{\text{max}}$ describes the maximum range of the non-quantized values. To determine it, we performed 10 rollouts with one of our models and set $x_{\text{max}}$ to the largest observed absolute value times $1.2$, per dimension. 

Note that explicit quantization would not be necessary in a real-world setup, as the sensor values are provided as integers to the controller.

The input normalization is implemented by offline computing new thresholds $\tau^*_{i,d}$, with $d$ being the dimension the threshold applies to and $i$ the index within the dimension. These are based on the original thresholds $\tau_{i,d}$, but account for both input normalization and quantization.
Then, with $\mu_{\text{running}}$ and $\sigma^2_{\text{running}}$ representing our frozen normalization statistics,

\begin{equation}
\label{eq:thresholds_star}
    \tau_{i,d}^* = \text{clip}\Bigg(\Bigg \lfloor \frac{\tau_{i,d} \sqrt{\sigma_{\text{running},d}^2} + \mu_{\text{running},d}}{Q_{S,d}} \Bigg \rfloor, -Q_{\text{max}}, Q_{\text{max}}\Bigg).
\end{equation}
This folds all multiplication constants into the thermometer thresholds, reducing the normalization and thermometer encoding to integer comparisons. Note that instead of the floor operation in Equation \ref{eq:thresholds_star}, the round-to-nearest operation $\lfloor \cdot \rceil$ could be used for computing the new thresholds. In our implementation, we use the floor operation in Equation~\ref{eq:thresholds_star}, and the rewards in Table~\ref{tab:end-to-end} are reported for this exact choice.

For the output mapping, we insert a single-port BRAM for each action dimension. We perform new evaluation rollouts with the now-quantized observations and recomputed thresholds, explaining minor differences in reward reported in Table \ref{tab:end-to-end}. 

Last, we provide an implementation report in the case where FP32 values are provided to the logic core and quantization handling is included in the reported resources (see Table \ref{tab:fp-included}). Note that this is an extremely unusual setup, as sensor values are not provided in FP32; we nevertheless provide these results for completeness. In this setting, we only provide results for HalfCheetah, Hopper, and Walker2d, as the other environments would require time multiplexing the available DSPs or selecting a different FPGA.

\section{Training Overhead}
\label{app:training overhead}
\begin{figure*}[t!]
\centering
\includegraphics{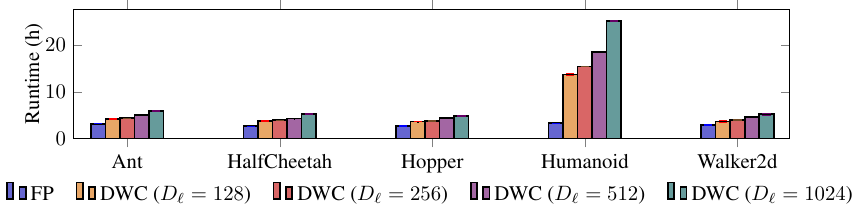}
\caption{Training runtime for one million environment steps, averaged over three training runs (mean $\pm$ std shown; standard deviation is so small that it is not visible). Floating-point models exhibit consistently shorter training times. However, it is only in the Humanoid case, which has a much larger number of observation dimensions than the other environments, that the training time for \acronyms is more than $\times3$ that of the floating-point baseline.}
\label{fig:runtime-comparison}
\end{figure*}

Figure \ref{fig:runtime-comparison} shows training times for different environments for one million training timesteps on an NVIDIA A10 GPU with an AMD EPYC 7513 32-Core Processor. %
We perform training on a GPU, as the current public implementation of DWNs only supports GPU inference and training. DWNs and LGNs are notoriously known for trading high efficiency at inference time \citep{petersen2022deep, bacellar2024differentiable} for increased training wall clock time. This is especially true when naive trainable interconnects are used \citep{kresse2025scalable}, as memory consumption and training time scale with the size of the connectivity matrix. This phenomenon is visible in our results: for almost all tasks, the DWCs required to solve the environments are comparatively modest in size with a small connectivity matrix. Consequently, training times are dominated by the cost of environment simulation steps rather than the network updates, yielding very similar overall times that remain within a $3\times$ factor of the floating-point network's training time up to $D_\ell = 1024$. 

The only exception is Humanoid, which exhibits significantly longer training times ($\times7.7$ when compared to the floating-point baseline for the largest network in Figure \ref{fig:runtime-comparison}). The longer training time is due to the much larger interconnect matrix between the first LUT layer and the observations, due to a much larger number of observations.

Future work should consider applying techniques introduced in the classification setting~\citep{kresse2025scalable,mommen2025method,fojcik2025lilogic,wang2026learning} that reduce the cost of the full-interconnect learning, while retaining most, if not all, performance benefits of the learnable interconnect.

\section{Hyperparameters}\label{app:hyperparams}

Table \ref{tab:sac_hparams} and Table \ref{tab:ddpg_hparams} show SAC and DDPG hyperparameters, respectively. Table \ref{tab:ppo_fp:hparams} shows the PPO floating-point hyperparameters used in our experiments, while Table \ref{tab:ppo_dwn:hparams} shows the PPO hyperparameters used for \acronyms. Hyperparameters for the LUT and interconnect training with EFD in the \acronyms are equivalent to the default configurations in \citet{bacellar2024differentiable}.

Major differences between FP and \acronyms PPO hyperparameters are a $\times10$ learning rate for \acronyms, a lower entropy coefficient, and a lower max gradient norm. The initial log $\alpha$ is conceptually similar to the floating-point std parameter, which corresponds to the initialization of the policy network's output variance. 

Tables \ref{tab:hyperparameter:tuning:bcc} shows the search space for the PPO hyperparameter tuning for \acronyms, while Table \ref{tab:hyperparameter:tuning:fp} shows the search space for the PPO hyperparameter tuning for the FP baseline. Remaining hyperparameters were set to the values in Table \ref{tab:ppo_fp:hparams}, which correspond to CleanRL \citep{huang2022cleanrl} defaults. Hyperparameter tuning was performed jointly across Ant, Hopper, Walker, and HalfCheetah, optimizing for return after 1M timesteps over 3 seeds. Hyperparameter search was performed with Optuna \citep{akiba2019optuna} using the Tree-Structured Parzen Estimator with 100 trials for both the floating-point and \acronym implementation. The hyperparameter search for \acronyms (PPO) took 25 days of compute, which we distributed over 10 GPUs.

\begin{table}[h!]
\centering
\small
\caption{SAC hyperparameters.}
\label{tab:sac_hparams}
\begin{tabular}{@{}ll@{}}
\toprule
\textbf{Hyperparameter} & \textbf{Value}  \\
\midrule
Total timesteps & $1{,}000{,}000$ \\
Replay buffer size & $1\times 10^{6}$ \\
Discount $\gamma$ & 0.99 \\
Target smoothing $\tau$ & 0.005 \\
Batch size & 256 \\
Learning starts & $5\times 10^{3}$ \\
Policy LR & $3\times 10^{-4}$ \\
Q-network LR & $1\times 10^{-3}$ \\
Policy update frequency & 2 \\
Target network frequency & 1 \\
Entropy & autotune \\
$\alpha_d$ (DWC only) & 0.5 \\
\bottomrule
\end{tabular}
\end{table}
\begin{table}[h!]
\centering
\small
\caption{DDPG hyperparameters.}
\label{tab:ddpg_hparams}
\begin{tabular}{@{}ll@{}}
\toprule
\textbf{Hyperparameter} & \textbf{Value} \\
\midrule
Total timesteps & $1{,}000{,}000$  \\
Learning rate & $3\times 10^{-4}$  \\
Replay buffer size & $1\times 10^{6}$  \\
Discount $\gamma$ & 0.99\\
Target smoothing $\tau$ & 0.005 \\
Batch size & 256  \\
Exploration noise (std) & 0.1 \\
Learning starts & $2.5\times 10^{4}$  \\
Policy update frequency & 2 \\
\bottomrule
\end{tabular}
\end{table}

\begin{table}[h!]
\centering
\small
\caption{PPO FP hyperparameters.}
\label{tab:ppo_fp:hparams}
\begin{tabular}{@{}ll@{}}
\toprule
\textbf{Hyperparameter} & \textbf{Value} \\
\midrule
Total timesteps & $1{,}000{,}000$ \\
Learning rate & $7.82 \times 10^{-4}$ \\
Number of environments & 1 \\
Steps per environment & 2048 \\
Number of minibatches & 32 \\
Update epochs & 10 \\
Discount $\gamma$ & 0.99 \\
GAE $\lambda$ & 0.95 \\
Clip coefficient & 0.2 \\
Entropy coefficient & $6.8 \times 10^{-5}$ \\
Value function coefficient & 0.5 \\
Max gradient norm & 0.811 \\
Floating-point std & 0.00212 \\
Adam $\epsilon$ & $1 \times 10^{-5}$ \\
Anneal LR & True \\
Norm advantages & True \\
Clip value loss & True \\
\bottomrule
\end{tabular}
\end{table}

\begin{table}[h!]
\centering
\small
\caption{PPO \acronyms hyperparameters. Not shown hyperparameters are equivalent to Table \ref{tab:ppo_fp:hparams}.}
\label{tab:ppo_dwn:hparams}
\begin{tabular}{@{}ll@{}}
\toprule
\textbf{Hyperparameter} & \textbf{Value} \\
\midrule
Learning rate & $6.76 \times 10^{-3}$ \\
Entropy coefficient & $3.37 \times 10^{-4}$ \\
Max gradient norm & 1.92 \\
Initial log $\alpha$ & -3.18 \\
LUT width & 256 \\
\bottomrule
\end{tabular}
\end{table}

\begin{table}[h!]
\centering
\small
\caption{Hyperparameter search space for \acronyms.}
\label{tab:hyperparameter:tuning:bcc}
\begin{tabular}{@{}ll@{}}
\toprule
\textbf{Hyperparameter} & \textbf{Search Space} \\
\midrule
Learning rate & Log-uniform $[7 \times 10^{-4}, 10^{-2}]$ \\
Max gradient norm & Uniform $[0.5, 2.5]$ \\
Hidden layer size & Categorical $\{128, 256, 512\}$ \\
Entropy coefficient & Uniform $[0, 0.05]$ \\
Initial log $\alpha$ & Uniform $[-6.0, -0.3]$ \\
\bottomrule
\end{tabular}
\end{table}

\begin{table}[h!]
\centering
\small
\caption{Hyperparameter search space for FP baseline.}
\label{tab:hyperparameter:tuning:fp}
\begin{tabular}{@{}ll@{}}
\toprule
\textbf{Hyperparameter} & \textbf{Search Space} \\
\midrule
Learning rate & Log-uniform $[7 \times 10^{-4}, 10^{-2}]$ \\
Max gradient norm & Uniform $[0.5, 2.5]$ \\
Entropy coefficient & Uniform $[0, 0.05]$ \\
Floating-point std & Uniform $[10^{-3}, 0.1]$ \\
\bottomrule
\end{tabular}
\end{table}

\section{LUT Input Ablation}

\begin{figure*}[t]
\centering
\includegraphics{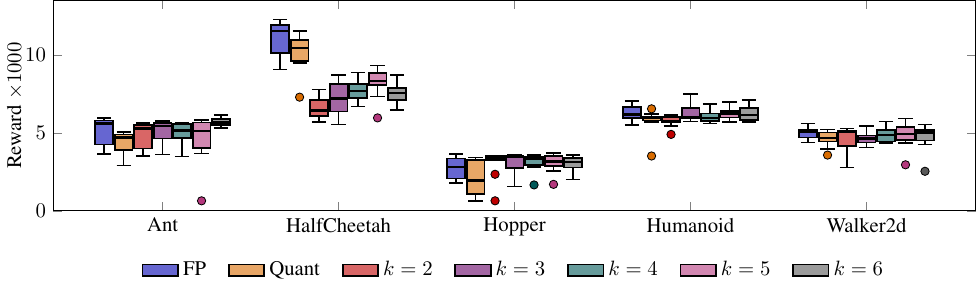}

\caption{Showing floating-point (FP) baseline returns across environments and \acronyms with varying LUT-inputs ($D_\ell=1024$). Generally, already models with 2-input LUTs achieve returns comparable to the FP baseline. Only for HalfCheetah, we observe a generally monotonically increasing median return with increasing LUT table size.}
\label{fig:k ablation}
\end{figure*}

Figure \ref{fig:k ablation} shows an ablation over the LUT input size \(k\) from 2 to 6 inputs. Larger \(k\) increases the expressivity of each LUT, but also exponentially increases the number of parameters per layer.

Clearly, already \(k=2\) achieves good results across all tasks, being comparable for all except for HalfCheetah. For HalfCheetah, we observed similar capacity-limited returns as we observed in our size ablation.

\section{Input Layer Bit \& Layer Number Ablation}\label{app:bits}
\begin{figure*}[t!]
\centering
\includegraphics{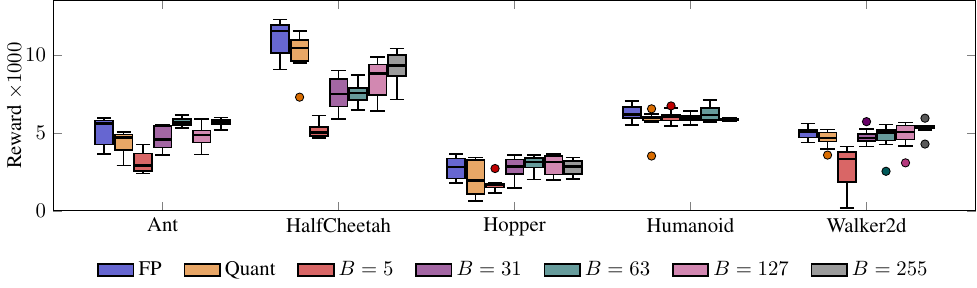}

\caption{Showing floating-point baseline and \acronyms return across enviroments with varying input bit width $B$. While having only 5 threshold bits per layer can harm policy performance. For most environments, 63 input bits suffice to saturate returns. 
As in other cases, the exception is HalfCheetah, which benefits monotonically from more fined-grained input binarizations.}
\label{fig:bit ablation}
\end{figure*}

\begin{figure*}[t!]
\centering
\includegraphics{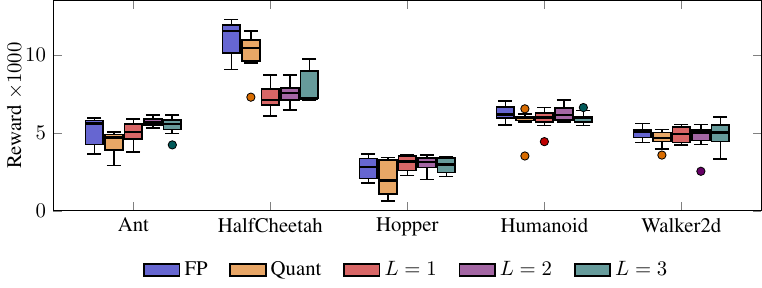}

\caption{Showing floating-point baseline and \acronyms return across environments with varying number of layers $L$. For $D_\ell$=1024, one layer generally suffices for attaining floating-point performance. 
Again, HalfCheetah, seems to benefit from additional layers, and hence, increased capacity.}
\label{fig:layer number ablation}
\end{figure*}

\begin{figure*}[t!]
\centering
\includegraphics{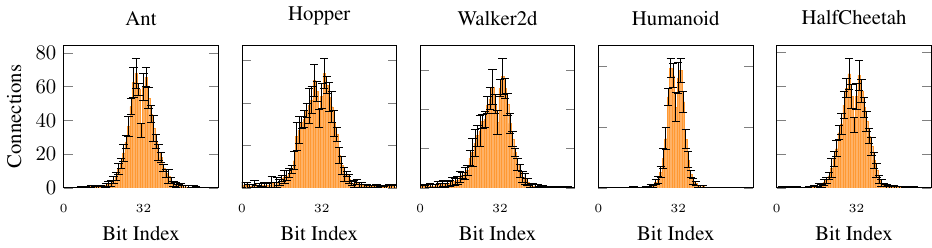}
\caption{Distribution of connections per Bit Index across all environments. $l=128$}
\label{fig:bitconnections:128}

\end{figure*}

Figure \ref{fig:bit ablation} shows an ablation over the number of thermometer bits used in the input layer. We observe that while performance suffers at 5 input threshold bits, policy performance does not completely collapse. For 63 bits and $l=128$, Figure \ref{fig:bitconnections:128} shows the learned per-bit index connectivity for all environments. Figure \ref{fig:layer number ablation} shows an ablation on the number of layers.

\section{XC7A15T FPGA resources}
\label{app:fpga}
Resources for the AMD Xilinx Artix-7 XC7A15T\textendash FGG484\textendash 1 device are shown in Table \ref{tab:artix}.
\begin{table}[h!]
  \centering
  \caption{XC7A15T\textendash FGG484\textendash 1 device resources.}
  \label{tab:artix}
  \begin{tabular}{l r}
    \toprule
    Resource & Quantity \\
    \midrule
    LUTs &10{,}400 \\
    Flip-flops & 20{,}800 \\
    DSPs & 45 \\
    Block RAM (36\,Kb) & 25 \\
    Max user I/O pins & 250 \\
    \bottomrule
  \end{tabular}
\end{table}

\end{document}